%% file: when-to-advise-arxiv-v2.tex
\documentclass[11pt,a4paper]{article}

\usepackage[left=0.9in, right=0.9in, top=1in, bottom=1in]{geometry}

\usepackage{amssymb}
\usepackage{amsmath}
\usepackage{amsthm}
\usepackage{amsfonts}
\usepackage{times}
\usepackage{soul}
\usepackage{url}
\usepackage[hidelinks]{hyperref}
\usepackage[utf8]{inputenc}
\usepackage[small]{caption}
\usepackage{graphicx}

\usepackage{booktabs}
\usepackage{algorithm}
\usepackage{algorithmic}
\usepackage[switch]{lineno}
\usepackage{varwidth}
\usepackage{placeins}

\usepackage{subcaption}
\usepackage{color}
\usepackage{makecell}
\usepackage{array}

\title{Learning When to Advise Human Decision Makers\footnote{This paper was published in Proceedings of IJCAI 2023. The data are available on the authors' websites. }}

\author{
Gali Noti\thanks{Harvard University and the Hebrew University of Jerusalem. Email: galinoti@seas.harvard.edu}
\and 
Yiling Chen\thanks{Harvard University. Email: yiling@seas.harvard.edu}
}

\date{}

\begin{document}

\maketitle

\begin{abstract}

\input{abstract-nature-ijcai}
\end{abstract}

\section{Introduction}

\input{intro-3-ijcai}

\section{Responsive Algorithmic Advising} 
\input{learning-sn-ijcai}

\section{Experimental Setup}
\input{setup-sn-ijcai}


\newcommand{\avgci}[2]{\makecell{{#1}\\{\scriptsize($\pm${#2})}}}

\begin{table*}
\small
\centering
\begin{tabular}{lccccc}
  & Learned & Random & Omniscient & No Advice & Update \\
	& N=218 & N=200 & N=200 & N=258 & N=220 \\
\midrule

Advising policy accuracy & \avgci{74.1\%}{0.9\%} & \avgci{58.4\%}{1.5\%} &  \avgci{100.0\%}{0.0\%} & \avgci{52.5\%}{2.0\%}  & \avgci{42.0\%}{2.1\%}  \\

Quadratic score & \avgci{0.781}{0.005} & \avgci{0.755}{0.007} & \avgci{0.825}{0.006} & \avgci{0.719}{0.008} & \avgci{0.770}{0.007} \\
Algorithm's quadratic score & \avgci{0.801}{0.003} & \avgci{0.800}{0.003} & \avgci{0.803}{0.003} & \avgci{0.805}{0.003} & \avgci{0.802}{0.003} \\
Linear score & \avgci{0.622}{0.006} & \avgci{0.578}{0.009} & \avgci{0.653}{0.007} & \avgci{0.560}{0.011} & \avgci{0.578}{0.008} \\
Algorithm's linear score & \avgci{0.603}{0.003} & \avgci{0.602}{0.004} & \avgci{0.604}{0.004} & \avgci{0.607}{0.003} & \avgci{0.603}{0.003} \\

Advice influence & \avgci{0.810}{0.036} & \avgci{0.769}{0.040} & \avgci{0.787}{0.041} & -- &  \avgci{0.321}{0.040} \\
Advice acceptance rates & \avgci{0.735}{0.043} & \avgci{0.683}{0.048} & \avgci{0.723}{0.045} & -- &  \avgci{0.305}{0.043} \\[0.4cm]

\makecell[l]{Human initial risk prediction that is at least as \\accurate as the algorithmic risk assessment} & \avgci{62.47\%}{1.66\%} & \avgci{56.84\%}{2.14\%} & \avgci{60.60\%}{1.89\%} & \avgci{52.46\%}{1.99\%} & \avgci{58.00\%}{2.06\%} \\[0.4cm]

\makecell[l]{KL divergence between the distributions of human initial \\risk prediction and the algorithmic risk assessment}  & 0.47 & 0.52 & 0.41 & 0.92 & 0.32  \\

False-positive rates (FPR) & \avgci{22.88\%}{1.82\%} & \avgci{39.39\%}{3.45\%} & \avgci{19.63\%}{2.30\%} & \avgci{52.64\%}{4.18\%} & \avgci{38.04\%}{3.39\%}  \\

False-negative rates (FNR) & \avgci{51.39\%}{1.99\%} & \avgci{43.55\%}{2.89\%} & \avgci{37.51\%}{2.72\%} & \avgci{36.28\%}{3.33\%} & \avgci{41.83\%}{2.62\%} \\

Classification disparity & 0.145 & 0.102 & 0.138 & 0.094 & 0.152 \\

\bottomrule
\end{tabular}
\caption{Experimental results: Overview of main metrics.}
\label{tbl:overview}
\end{table*}

\section{Results}

\input{results-ijcai} 

\section{Discussion}
\input{discussion-ijcai}

\section*{Ethics Statement}
\input{ethical_statement}

\section*{Acknowledgments}
This project is partially supported by U.S. National Science Foundation under grant No. IIS 2007887 and grant No. IIS 2147187.
The project has also received funding from the European Research Council (ERC) under the European Union’s Horizon 2020 research and innovation programme
(grant agreement No 740282).

\bibliographystyle{abbrv}
\bibliography{when-to-advise-less-info}

\newpage

\appendix

\section*{APPENDICES}

\section{Further Related Work} 
\input{related-work} \label{app:further-related-work}

\section{The Pretrial-Release Decision Setting and Dataset} \label{app:setting}
\input{setting}

\section{Additional Details on Learning the Advising Policy} \label{app:learning}

\input{app-results}

\end{document}

%% file: abstract-nature-ijcai.tex
Artificial intelligence (AI) systems are increasingly used for providing advice to facilitate human decision making in a wide range of domains, such as healthcare, criminal justice, and finance. Motivated by limitations of the current practice where algorithmic advice is provided to human users as a constant element in the decision-making pipeline, in this paper we raise the question of {\em when should algorithms provide advice?} We propose a novel design of AI systems in which the algorithm interacts with the human user in a two-sided manner and aims to provide advice only when it is likely to be beneficial for the user in making their decision. The results of a large-scale experiment show that our advising approach manages to provide advice at times of need and to significantly improve human decision making compared to fixed, non-interactive, advising approaches. This approach has additional advantages in facilitating human learning, preserving complementary strengths of human decision makers, and leading to more positive responsiveness to the advice. 

%% file: intro-3-ijcai.tex
Artificial intelligence (AI) is increasingly used to support human decision making in high-stake settings in which the human operator, rather than the AI algorithm, needs to make the final decision.
For example, in the criminal justice system,  algorithmic risk assessments are being used to assist judges in making pretrial-release decisions and at sentencing and parole \cite{courts2020,pji2019,compas2019,cohen2018}; in healthcare, AI algorithms are being used to assist physicians to assess patients' risk factors and to target health inspections and treatments \cite{musen2021,garcia2019,tomavsev2019,kononenko2001}; and in human services, AI algorithms are being used to predict which children are at risk of abuse or neglect, in order to assist decisions made by child-protection staff \cite{vaithianathan2017,chouldechova2018}.

In such systems, decisions are often based on 
risk assessments, 
and statistical machine-learning 
algorithms' abilities to excel at prediction tasks \cite{meehl1954,dawes1989,grove2000,obermeyer2016,mullainathan2017} are leveraged to provide predictions as advice to human decision makers \cite{kleinberg2015}. 
For example, the decision 
that judges make
on whether it is safe to release a defendant until his trial, is based on their assessment of how likely this defendant is, if released, to violate his release terms, i.e., to commit another crime until his trial or to fail to appear in court for his trial. For making such risk predictions, judges in the US are assisted by a ``risk score'' predicted for the defendant by a machine-learning algorithm \cite{courts2020,pji2019}. 

Research on such AI-assisted decision making has mostly 
addressed two questions. The first is what advice should AI systems provide? The line of research that addresses this question places emphasis on the machine-learning algorithms and focuses on optimizing 
and evaluating their success in comparison to human predictions, based on statistical metrics 
such as prediction accuracy and fairness \cite{kleinberg2017,angwin2016,haenssle2018,chouldechova2017}. 
The implicit expectation is that better algorithmic advice will lead to better human decisions. 

The second question is how to present algorithmic advice to human decision makers? This question has been addressed in a recent line of work that emphasizes the role of the human as the one who eventually makes the actual decision. Instead of evaluating the algorithmic performance in isolation, these works concentrate on studying the effect of the algorithmic input on the decisions that humans make  \cite{benFAT,benCSCW,albright2019,tschandl2020,lai2019,zhang2020,bansal2019,yin2019} and hence term the perspective ``AI-in-the-loop human decision making''\cite{benFAT}. These studies typically show---both with human experts such as judges or clinicians and with non-experts in experimental settings---that providing the algorithmic assessment indeed significantly improves human decision makers' prediction performance, and that different ways of providing the algorithmic input to human decision makers, as well as different algorithmic accuracy or error patterns, can have a significant impact on their decisions.

Situated in the framework of AI-in-the-loop human decision making, this work aims to answer a different important question:  {\em when should algorithms provide advice?} The current practice in applications and in prior studies, is that algorithms provide advice to the human decision maker in every prediction problem. 
We explore whether AI systems can be trained to automatically identify the cases where advice is most useful, and those where the human decision maker is better off deciding without any algorithmic input, and whether such an approach that provides the algorithmic advice only when it is needed indeed manages to 
assist humans in improving their decisions.

\subsection{Background and Related Work}

Our approach is motivated by several observations from prior work and current practice.  
First, in  
prior studies on AI-assisted human decision making, the AI component is completely oblivious of the human decision maker: the human always receives advice from an algorithm, 
but, importantly, the algorithm is not aware of its human counterpart and whether its advice may actually be helpful to him. This is despite the fact that human decision makers have their own strengths and 
sometimes reach better decisions on their own, 
without the algorithmic input, and the computational methods have their own limitations and can have errors and biases (as was studied in recent literature on human-AI complementary performance \cite{groh2022,wilder2020,madras2018,kamar2012,bansal2021,steyvers2022}), and so algorithmic advice may not always be helpful.

E.g., 
in a recent experiment that studied human prediction in the pretrial-release decision setting \cite{benCSCW}, 
the most accurate human predictions were achieved in an ``Update'' treatment, 
in which the human decision makers first made a risk prediction on their own and only then observed the algorithmic prediction and were allowed to update their prediction if they wished. 
However, in this dataset we found that in 66\% of the predictions, the human's initial prediction (before observing the algorithmic input) was already equal to or more accurate than the algorithm's prediction. Moreover, in 36\% of the predictions, humans'  initial prediction was strictly more accurate than the algorithm's, and after showing them 
the algorithmic prediction their prediction performance deteriorated 32\% of these times.

An additional important point that arises when a human decision maker is assisted by an (inevitably) imperfect AI system, is that the human is de-facto expected to monitor the algorithm, i.e., to identify when the algorithm is wrong so as to override its prediction \cite{ben2021}. However, there is a large body of empirical evidence showing that such monitoring is a challenging task for humans: recent studies demonstrate that people do poorly in judging the quality of algorithmic predictions and determining when to override those predictions, and that these judgments are often incorrect and biased \cite{benFAT,benCSCW,grgic2019,tschandl2020,yeomans2019,bansal2021,swol2005}. 
This suggests to consider alternative designs of decision pipelines in which the monitoring task 
is transferred to the AI.

Moreover, even if the algorithm were perfect, it is not clear whether the constant advising approach used in prior work is the optimal way to interact with human decision makers and to inform them so as to improve their decisions. Specifically, it may be that providing the advice in every prediction will result in advice discounting or even disregard in the decision maker's judgment.  
Such behaviors have been demonstrated in other settings of users' interactions with technology \cite{kalsher2006,anderson2014}, 
and are related to the study of habituation \cite{rankin2009},  
but have not been studied in behavioral literature on advice utilization.

Finally, in the experiment of \cite{benCSCW} mentioned above, it is intriguing to see that while humans made significantly better predictions when they were assisted by the algorithm compared to making predictions without any algorithmic assistance, their performance was still far worse than that of the algorithm alone. This is despite the fact that the human decision makers constantly received the algorithmic prediction, and, in principle, could just have adopted its predictions and reached the algorithmic performance. 
This observation that a human assisted by an algorithm is still inferior to the algorithm alone is in fact typical in AI-assisted human decision making settings (e.g., \cite{lai2019,lai2020,jacobs2021}) 
and suggests that there is room to improve and extract more value from the interaction between the human and the AI.
For a discussion of further related work, see Appendix \ref{app:further-related-work}.

\begin{figure*}[!t]
\centering
\begin{subfigure}{.49\linewidth}
\center
\includegraphics[width=1.0\linewidth]{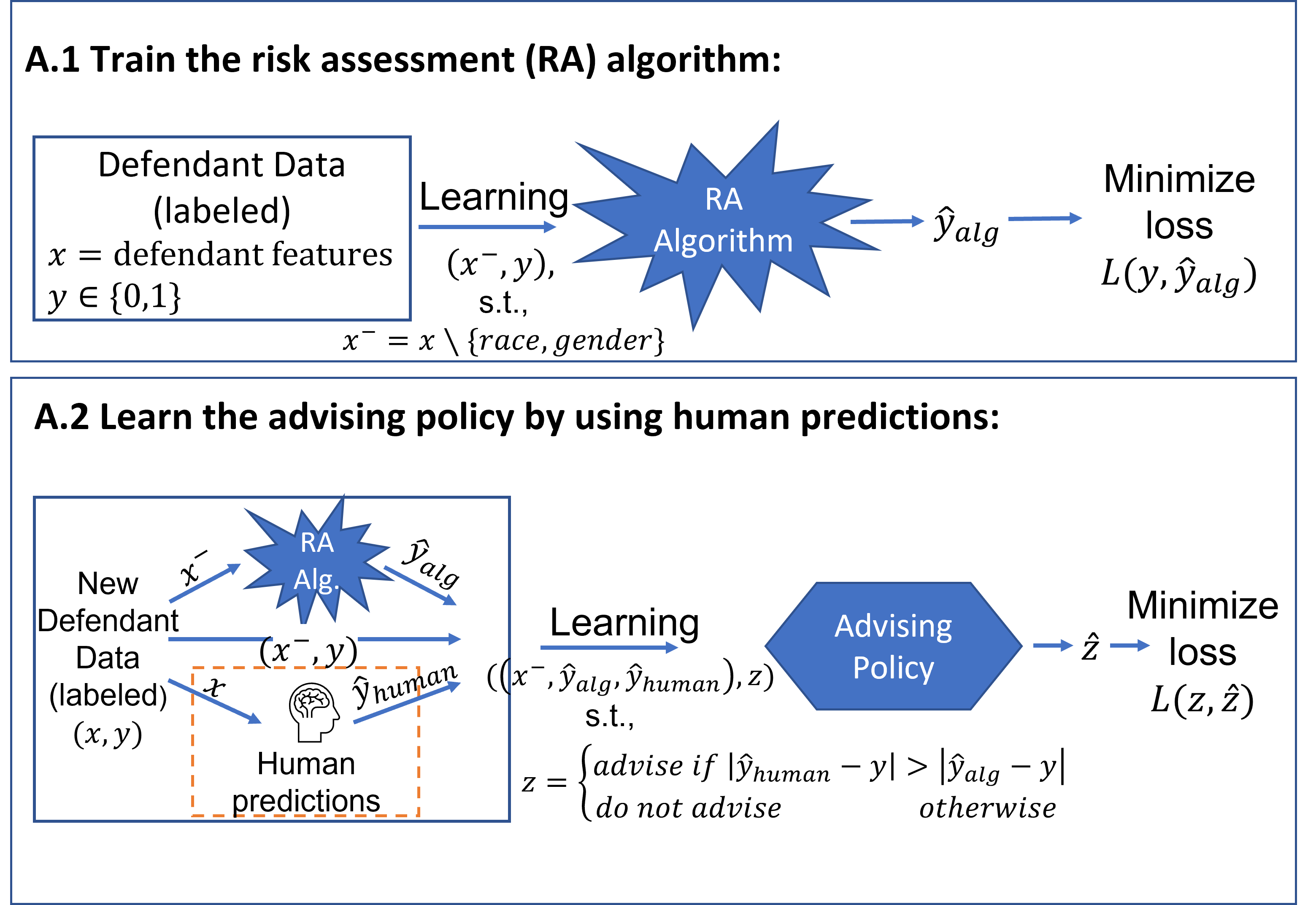}
\caption{Training the algorithmic components}
\label{fig:framework-a}
\end{subfigure} 
\begin{subfigure}{.49\linewidth}
\center
\includegraphics[width=1.0\linewidth]{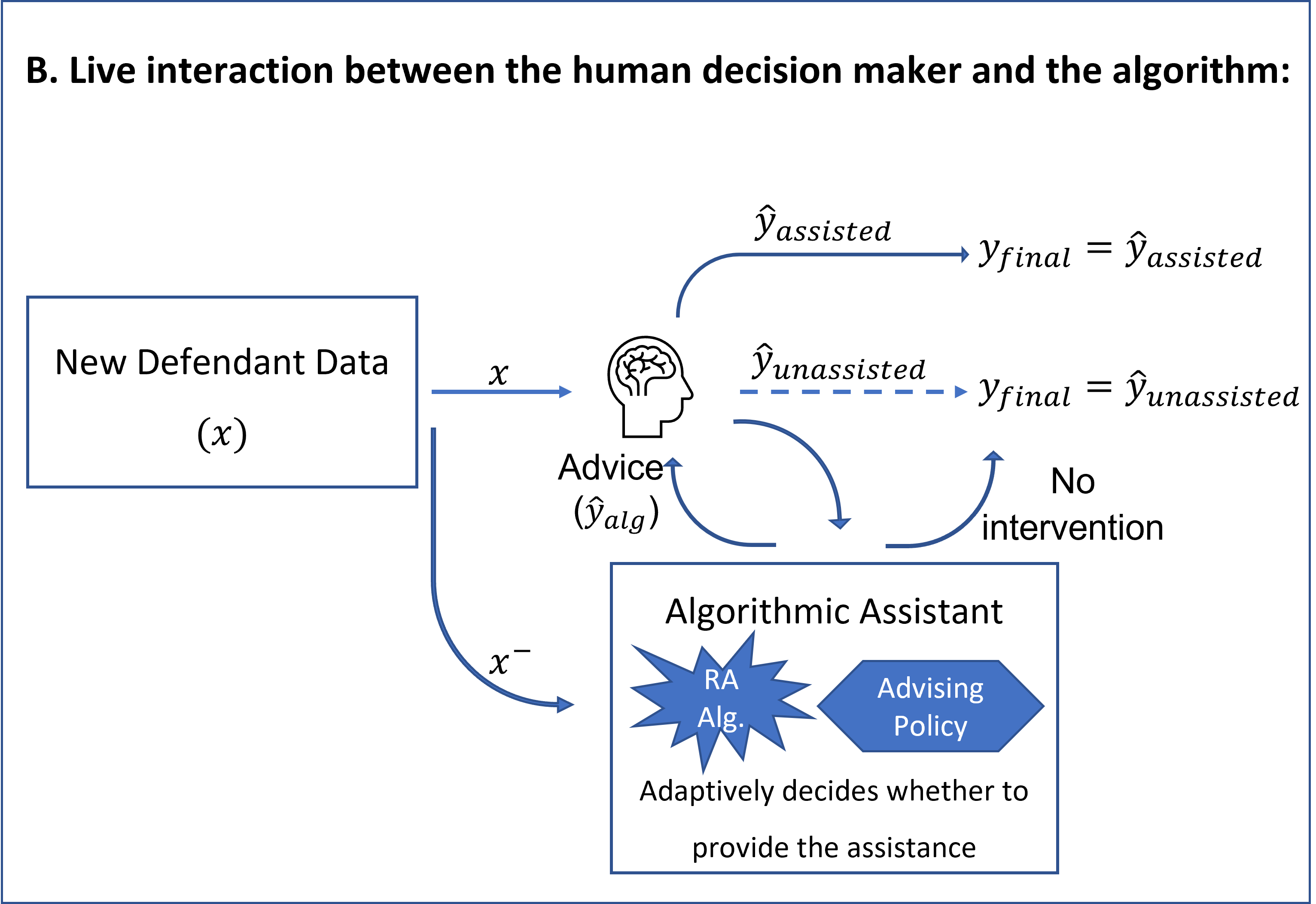}
\caption{Live interaction}
\label{fig:framework-c}
\end{subfigure}
\caption{{\small  A responsive advising  approach for AI-assisted human decision making.}}
\label{fig:framework}
\end{figure*}

\subsection{Our Approach}

We propose to replace the constant advising approach with 
a responsive advising system (an ``algorithmic assistant'') 
that interacts with the human decision maker and takes an active part in the decision making process, aiming to improve the human's decisions. Specifically, our algorithmic assistant applies a {\em learned advising policy} that depends on input from the human decision maker and provides advice only when it is likely to improve his decision. Thus, in this human-AI team, information does not only flow from the algorithm to the human as in prior work, but instead there is a \emph{two-sided interaction}: 
the algorithmic assistant's advice   
depends on 
the human's input, and the human's final decision,  
in turn, depends on 
the input he receives from his algorithmic assistant.

We consider a simple form of these two-sided interactions, in which the input from the human to the algorithmic assistant is the human's (initial, unassisted) risk prediction, and the algorithmic assistant's advising policy determines whether or not to advise the human, providing advice only when it identifies that its advice is likely 
to improve the human's prediction. 
Thus, our human-AI collaboration is designed such that the human decision maker is operating on his own and makes predictions, while the learned advising policy is there to optimize the added value that the human can extract from his interaction with the AI advising system.

Figure \ref{fig:framework} presents a diagram of the advising-policy 
approach that we take for AI-assisted decision making, which we demonstrate in the pretrial-release decision setting. 
We first learn an advising policy by using predictions that humans made in previous experiments about how likely a criminal defendant is to violate his release terms if released. Then, we conduct a large-scale experiment on Mechanical Turk to evaluate human prediction performance when 
decision makers interact with 
our learned advising policy. Our experimental results show that an advising policy is indeed learnable from data and that humans assisted by this learned advising policy make significantly more accurate predictions than human decision makers assisted by the constant advising policy, and achieve comparable performance to the risk-assessment algorithm, thus improving over the state-of-the-art. 
We further explore how our responsive-advising approach affects 
human learning, 
human decision makers' responsiveness to algorithmic advice,   
and 
the performance of  
human decision makers 
with respect to defendants' racial groups.

%% file: learning-sn-ijcai.tex
The decision pipeline that we consider is illustrated in Figure \ref{fig:framework-c}. It is composed of 
a human decision maker and an algorithmic assistant. The algorithmic assistant is in turn composed of a risk-assessment algorithm and an advising policy.\footnote{Note that in principle, an algorithmic assistant could be implemented as a single  algorithmic  
component trained end-to-end. However, such an approach would not allow us to isolate the contribution of the advising policy to  
human prediction performance. Furthermore, an important advantage of decoupling the risk-assessment algorithm from the advising policy is in reliability: such a separation constrains the risk-assessment algorithm to be trained only to optimize the quality of its risk assessments, rather than providing biased assessments that aim to affect the human decision maker. 
} 
When a new criminal defendant arrives, the human observes a description of the defendant and predicts the defendant's likelihood to violate his release terms if released, $\hat{y}_{unassisted}$. Then, given the description of the defendant (excluding race and gender to match common practice among risk-assessment developers and previous experiments \cite{laura2016,lowenkamp2009,cohen2018,benFAT,benCSCW}) and the prediction that the human made, the algorithmic assistant generates an algorithmic risk assessment, $\hat{y}_{alg}$, and provides the assessment to the human according to the advising policy.
The advising policy that we wish to learn aims to provide the algorithmic risk assessment to the human 
only when it is likely to improve the human's prediction. 
In cases in which the advice is not provided, the final prediction $\hat{y}_{final}$ is set to the human's unassisted prediction $\hat{y}_{unassisted}$, while in cases that the advice is provided, the human 
observes the advice and can
update his prediction if he wishes (to any prediction value), and the final prediction is then set to the updated value $\hat{y}_{assisted}$.

Our main focus is on learning such advising policies and evaluating their impact on the predictions made by human decision makers. As the risk-assessment component of the algorithmic assistant, we use the model of \cite{benCSCW} that was trained on 47,141 defendant cases from a dataset collected by the U.S. Department of Justice \cite{data2014} (top diagram in Figure \ref{fig:framework-a}), all of which were released before trial and we thus know the ground-truth information about their pretrial-release outcome. That is, we know for each case whether eventually the defendant violated his release terms. Among the defendants in this dataset, 29.8\% violated their pretrial-release terms. The model gets as input a description of a criminal defendant  
and outputs a risk assessment $\hat{y}_{alg}\in[0,1]$ that represents the algorithm's predicted likelihood that this defendant will violate his release terms if released. In \cite{benCSCW} it is shown that this model achieves comparable performance to widely used risk-assessment tools like COMPAS \cite{compas2017} 
and the Public Safety Assessment \cite{desmarais2016}. See  Appendix \ref{app:setting}  
for more details on the dataset and the model.

For learning the advising policy, we train a 
random-forest 
model on experimental data of human predictions from \cite{benFAT}. See the bottom diagram in Figure \ref{fig:framework-a}.
Given a defendant case, the algorithmic risk assessment, and the prediction that the human made, the policy determines whether or not to advise the human. 
In the training process, the label of each such a prediction example is set to 1 (i.e., do advise) if the algorithm's prediction is more accurate than that made by the human, and to 0 (i.e., do not advise) otherwise. 
The training data consist of 6,250 predictions made by 250 human participants, 
for 500 defendant cases.
Each participant was asked to predict for a series of 25 defendants, the defendants' risk 
to violate their release terms if released. In this dataset, in 33.31\% of the predictions the algorithm's prediction was more accurate than the human's prediction.
To better adapt to our target domain, which is a new experiment in which humans interact with a learned advising policy rather than predict independently from it as in our training data, we train our model on an augmented version of the dataset. 
For more details on the learning process, see Appendix \ref{app:learning}.

%% file: setup-sn-ijcai.tex
We conducted an experiment on Mechanical Turk to evaluate the quality of human predictions when assisted by our learned advising policy. In the experiment, each participant was randomly assigned to one of the experimental treatments (see below), and was asked to predict the risk for a series of 50 defendants 
(from 0\% to 100\%, in 10\% intervals) to violate their release terms if released, according to the decision pipeline described above. 
Overall, there were 1,096 participants in the experiment, who made a total of 54,800 predictions. The experimental data are available on the authors' website. 

Our experimental design compares human prediction performance in five experimental treatments. The first three treatments compare human performance when assisted by advising policies of different learning quality: \textbf{``Learned,''} in which humans were assisted by the learned advising policy described above; \textbf{``Random,''} in which the subset of defendant cases for which the human received the algorithmic advice was chosen at random, 
in the same frequency in which the learned advising policy provided advice on the training data; 
\textbf{``Omniscient,''} in which humans were assisted by an advising policy that showed the advice exactly in those cases where the algorithmic risk assessment was more accurate than their initial (unassisted) prediction, based on the ground truth of the defendant case (i.e., whether the defendant eventually violated his release terms). 
This provides an upper bound for performance improvement that may be achieved by improving the learning quality of our advising policy. 

In addition, we ran a \textbf{``No Advice''} treatment in which humans made the predictions on their own without observing the algorithmic risk assessment, and the \textbf{``Update''} treatment from \cite{benCSCW}, in which humans first made the prediction on their own and then always observed the algorithmic prediction and were allowed to update their prediction if they wished. The prediction structure in this Update treatment led to the best human prediction performance in \cite{benCSCW}, consistently with findings in other recent studies (e.g., \cite{bucinca2021,groh2022}) and with prior behavioral research that suggest the importance of forming a pre-advice independent opinion  \cite{swol2005,bonaccio2006,sniezek1995}.

For comparability with the experimental results of \cite{benCSCW}, we used in the experiment the same set of 300 defendant cases that they used, which were sampled from the heldout dataset of the risk-assessment algorithm's training process, and followed their experimental setup and procedure. 200 people or more participated in each of the experimental treatments.  
For more details, see Appendix \ref{app:setup}.

%% file: results-ijcai.tex
Next, we describe the main experimental results. Table \ref{tbl:overview} 
provides an overview of the results according to the main metrics.  All p-values and confidence intervals are generated on distribution of performance at the participant level, unless otherwise stated.

\subsection{Learning Performance} 
Our analysis starts by evaluating the extent to which our learned advising policy managed to generalize from the fixed training data to the new domain of our experiment, which includes new participants, new defendant cases, and importantly, live interaction between the advising policy and the human decision maker. 
The experimental results show that our learned advising policy managed to provide the advice in the correct times, i.e., when the algorithmic risk assessment was more accurate than the human's initial risk prediction, significantly more frequently than all other treatments (except, of course, from the Omniscient treatment, which by definition has perfect accuracy), obtaining accuracy of 74.1$\pm$0.9\%. 
This is compared with 58.4$\pm$1.5\% accuracy in the Random treatment in which the advice is given at random times;  with 42.0$\pm$2.1\% accuracy in the Update treatment which can be thought of as an ``always advising policy;'' and with 52.5$\pm$2.0\% accuracy in the No Advice treatment which can be thought of as a ``never advising policy.'' In the Learned treatment, in 37.5\% of the predictions the algorithmic risk assessment was more accurate than the human's initial risk prediction, and our learned advising policy provided the advice in 37.0\% of the predictions, thus achieving calibrated advice frequency. For further details, see Appendix \ref{app:results}. 

\subsection{Impact on Human Prediction Performance} 

\begin{figure}
	\centering
		\includegraphics[width=1.0\linewidth]{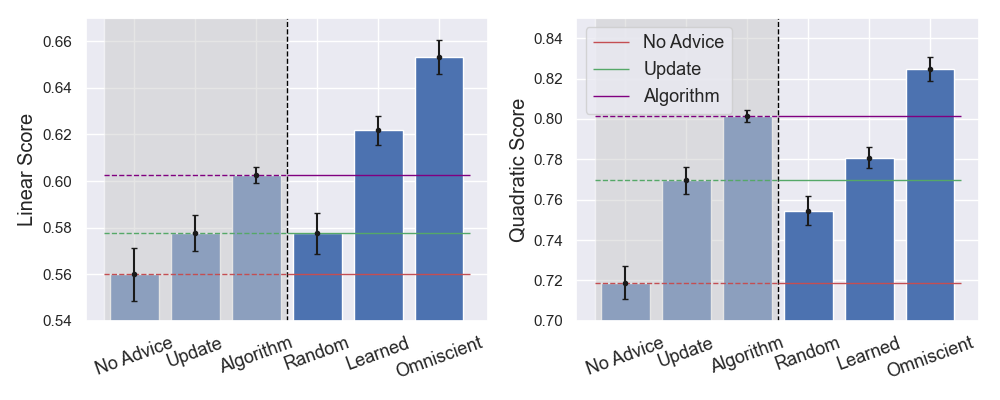}
	\caption{{\small Participant performance in the different experimental treatments, and the algorithm's performance, according to both linear and quadratic scores. Error bars show 95\% confidence intervals. The No Advice, Update, and algorithmic benchmarks are presented on the left of each figure, and for ease of comparison, their means continue as horizontal lines. 
The algorithm's performance is computed over its predictions for the cases given to participants in the Learned treatment.}}
\label{fig:scores-ci}
\end{figure}

We turn to look at the actual impact of our learned advising policy on the quality of the final predictions of the human decision makers. For each risk prediction $\hat{y}\in\{0,0.1,...,1\}$ 
with ground truth $y\in\{0,1\}$ (0 for not violating the release terms or 1 otherwise), the prediction error is defined as $error=\lvert y-\hat{y} \rvert$. We evaluate the prediction performance primarily according to two measures that capture different error patterns: the linear score (i.e., $1-error$) and the quadratic score (i.e., $1-error^2$), which is a proper scoring rule \cite{gneiting2007}. Evaluation according to additional measures gives qualitatively similar results (see Appendix \ref{app:results-impact}). 

Figure \ref{fig:scores-ci} shows the prediction performance of the human participants in the experiment and the algorithmic prediction performance, according to the linear score (left panel) and quadratic score (right panel). 
See Appendix \ref{app:results-impact} for the
 full performance distributions.
According to both score measures, human predictions in the Learned treatment have a clear and statistically significant advantage over the No Advice, Random, and Update treatments, and specifically the ranking of performance, 
from best to worst, is: Omniscient, Learned, Update, Random, and No Advice. 
The advantage of the Learned treatment over the constant-advising Update treatment demonstrates the usefulness of our learned advising policy approach that considers input from the human decision maker, and provides advice that is focused only on  those places where it is likely to be useful. The performance of the Random treatment shows that providing advice only in part of the predictions does not lead in itself to an improvement in the quality of human predictions, and that  
the learned advising approach 
is important to achieve this improvement. 
The large gap of the performance of Omniscient above all other treatments shows the potential for further improvement of human predictions by improving the learning quality of the advising policy (e.g., by utilizing more advanced computational methods or larger datasets).

A comparison with the algorithmic performance shows that according to the linear score human decision makers in the Learned treatment outperformed the algorithm, while according to the quadratic score the algorithm had better performance.\footnote{Note that the algorithm we use was trained to optimize quadratic score, and thus it could be expected that it will have an advantage according to this measure compared to other measures.}
Thus, we conclude that the prediction performance of human decision makers when assisted by our learned advising policy was on par with the performance of the algorithm.
Humans in the Omniscient treatment outperformed the algorithm by a large gap according to both measures, which again shows the potential for further gains from improving the learning quality. 
Reaching the algorithmic performance is a notable improvement compared to prior advising methods in the human-AI collaboration in decision making setting that we consider that requires human agency, and in particular compared with the constant-advising Update treatment, in which human prediction performance is typically significantly inferior to that of the algorithm.


\subsection{Interaction between the Human Decision Maker and the Algorithm}

\begin{figure}[!t]
\centering
\begin{subfigure}{0.4\linewidth}
\center
\includegraphics[width=1.03\linewidth]{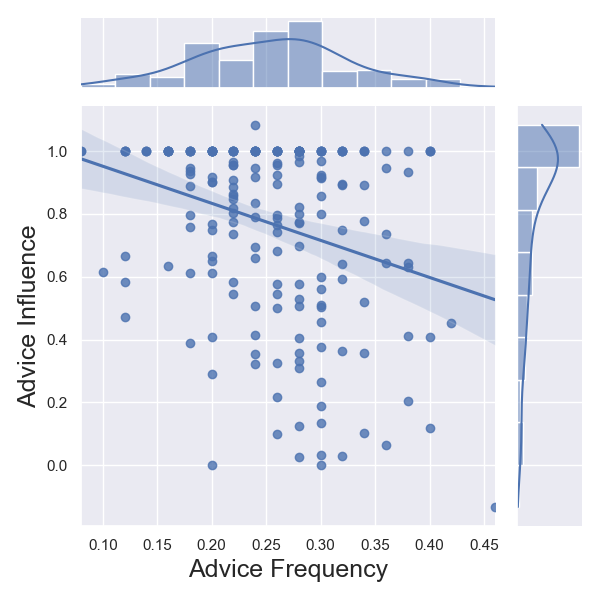}
\caption{} 
\label{fig:scarcity-effect-influence}
\end{subfigure}
\begin{subfigure}{0.4\linewidth}
\center
\includegraphics[width=1.03\linewidth]{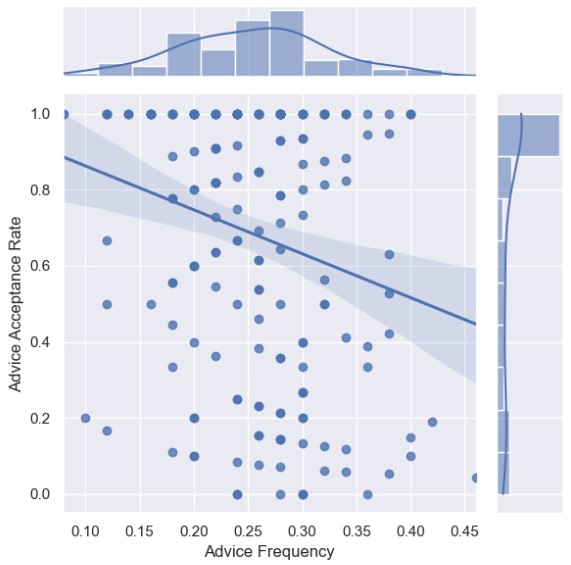}
\caption{} 
\label{fig:scarcity-effect-acceptance-rate}
\end{subfigure}
\caption{The scarcity effect. The figures show for participants in the Random treatment, scatter plots of the advice influence (Figure \ref{fig:scarcity-effect-influence}) and the advice acceptance rate (Figure \ref{fig:scarcity-effect-acceptance-rate}) vs. the advice frequency, alongside the marginal distributions and the regression lines.}
\label{fig:scarcity-effect}
\end{figure}

\paragraph{Human Responsiveness to the Algorithmic Advice.}
We now look at the responses of the human decision makers 
to the algorithmic advice. 
We measure human responsiveness to the algorithmic advice, in those cases in which the advice was given, by two measures: 
\begin{enumerate}
	\item {\em Advice influence} \cite{benCSCW}: for each prediction $\hat{y}^k_{unassisted}$ for which an advice $\hat{y}^k_{alg}$ was given, the influence is defined by $I^k = (\hat{y}^k_{assisted} - \hat{y}^k_{unassisted}) / (\hat{y}^k_{alg} - \hat{y}^k_{unassisted})$. This measure quantifies the extent to which the human prediction after observing the advice changed from its initial value in the direction of the value of the advice.
It is similar to the ``weight of advice'' measure \cite{yaniv2004}: when the final (assisted) prediction falls within the initial (unassisted) prediction and the advice, the influence reflects the weight that a participant assigns to the advice. Influence of 0 means that the participant ignored the advice, while an influence of 1 means that the participant adopted the advice exactly. 
The influence values ranged in $[-6,5]$, with 86.2\% of the predictions in $[0,1]$. 
\item {\em Advice acceptance rate}: considering predictions where the initial 
prediction is different than the algorithmic risk assessment, the advice acceptance rate is the frequency in which the advice is exactly followed. I.e., $Pr(\hat{y}^k_{assisted} = \hat{y}^k_{alg} \lvert \hat{y}^k_{unassisted} \neq \hat{y}^k_{alg} \text{ and } \hat{z} = 1)$.
\end{enumerate}

We observe a clear pattern (Figure \ref{fig:scarcity-effect}), which we term a ``scarcity effect'': 
as the advice is given less frequently, it tends to be followed by a stronger response on the human decision maker's part. 
Specifically, we look at the Random treatment, in which advice is given at random times and thus there is a natural variance in the frequencies in which participants received the advice. 
We find that the advice frequency is negatively correlated with the responsiveness of participants to the advice, as measured by the advice acceptance rate ($\rho=-0.23$, $p<0.001$) and by the advice influence measure 
($\rho=-0.29$, $p<0.0001$). 
Additionally, we observed that human responsiveness to the advice in the partial-advising treatments (Random, Learned, and Omniscient) was stronger, by a large gap, than the responsiveness in the Update treatment in which algorithmic risk assessment was provided for all predictions (Figure \ref{fig:responsiveness} in Appendix \ref{app:results-responsiveness}).  
While the scarcity effect we observed in the Random treatment is sufficiently strong to explain such a gap (by extrapolating the correlation pattern to an advice frequency of 100\%), this gap could also result from  
other factors, 
and 
our experimental design does not isolate the sources for the gap in human responses between these treatments.  See 
Appendix \ref{app:results-responsiveness}  
for more details.

\paragraph{Indication of Human Learning.}

In order to see whether our human decision makers managed to learn and improve over the course of the experiment, we analyze the quality of participants' initial (i.e., unassisted) prediction in comparison with the algorithm's prediction (which is the only type of feedback that the participants received in the experiment).  
Note that in all treatments participants had the same information when making their initial predictions, and so  differences between treatments in the quality of these predictions are a result of some learning process from the interaction with the different advising policies. 

The results show that in all experimental treatments participants managed to learn and improve their initial predictions relative to the No Advice benchmark, and suggest that the informed advising policies, namely Learned and Omniscient, better facilitate human learning. 
Specifically, our first indication of human learning is that the overall frequency in which the human initial prediction was at least as accurate as that of the algorithm, was significantly higher than No Advice in all experimental treatments, and was the highest in  
the Learned and Omniscient treatments (Figure \ref{fig:learning-overall} in Appendix \ref{app:results-responsiveness}). 
Second, looking over time, we find that this frequency significantly increased with prediction period only in the Learned and Omniscient treatments (Figure \ref{fig:learning-over-time} and analysis in 
Appendix \ref{app:results-responsiveness}). 
While these observations show a clear learning effect with respect to the algorithmic feedback that participants received, we find that this effect was only weakly translated to an improvement in the quality of the initial predictions with respect to the ground truth. 
See Appendix \ref{app:results-responsiveness} 
for further details. 

\begin{figure*}[!t]
\centering
\begin{subfigure}{.19\linewidth}
\center
\includegraphics[width=1.0\linewidth]{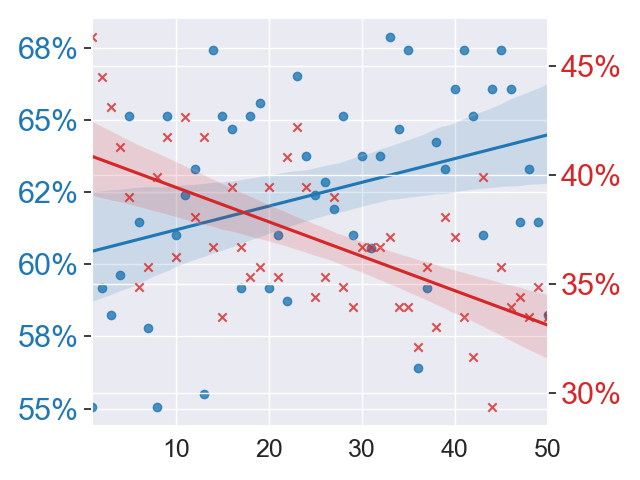}
\caption{Learned}
\label{fig:wheels-learned}
\end{subfigure}
\begin{subfigure}{.19\linewidth}
\center
\includegraphics[width=1.0\linewidth]{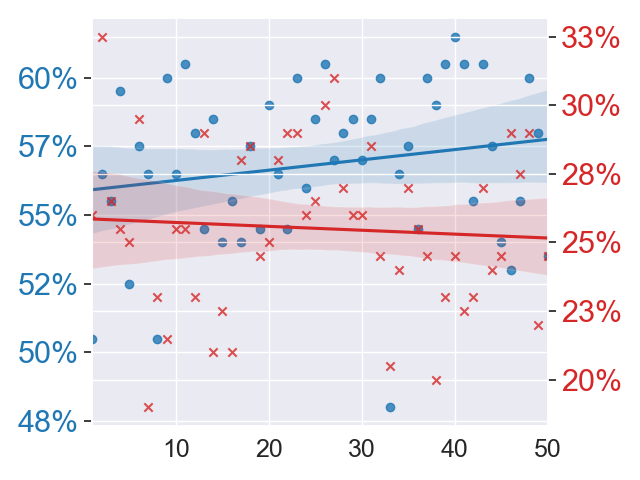}
\caption{Random}
\label{fig:wheels-random}
\end{subfigure}
\begin{subfigure}{.19\linewidth}
\center
\includegraphics[width=1.0\linewidth]{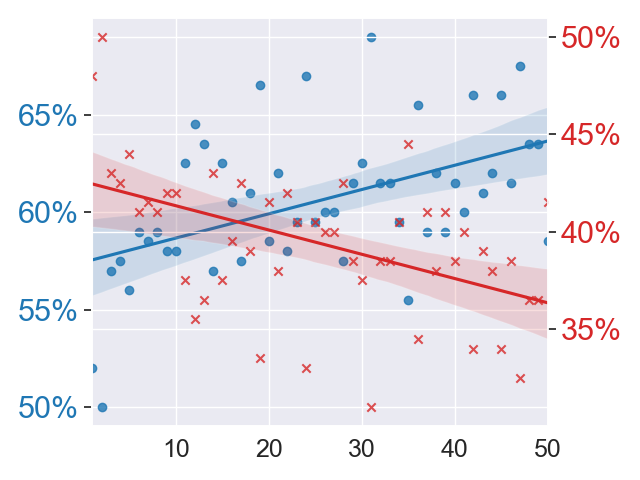}
\caption{Omniscient}
\label{fig:wheels-oracle}
\end{subfigure}
\begin{subfigure}{.19\linewidth}
\center
\includegraphics[width=1.0\linewidth]{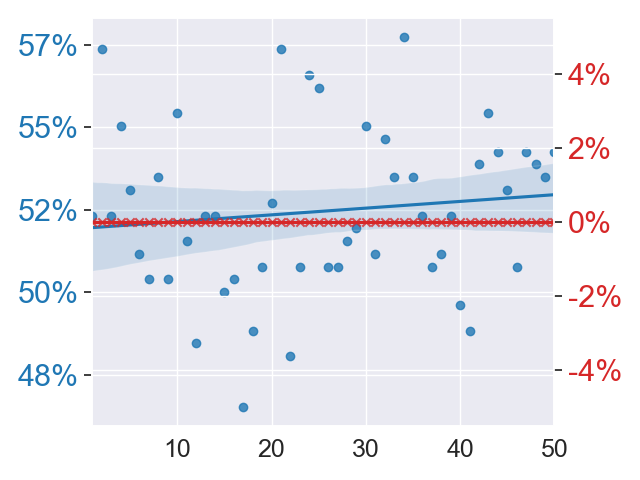}
\caption{No Advice}
\label{fig:wheels-baseline}
\end{subfigure}
\begin{subfigure}{.19\linewidth}
\includegraphics[width=1.0\linewidth]{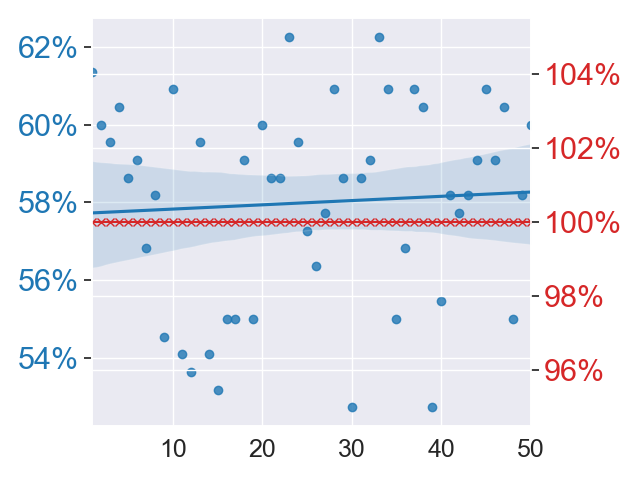}
\caption{Update}  
\label{fig:wheels-update}
\end{subfigure}
\vspace{-2pt}
\caption{{\small Human learning. In blue: the average frequency in which the human initial prediction was at least as accurate as the algorithm's prediction, as a function of the prediction period. In red: the average frequency in which the advice was provided as a function of the prediction period.
The blue and red lines are the regression curves for the two measures. Only in the Learned and Omniscient treatments these frequencies are significantly correlated with prediction period, and this learning effect resembles a ``training wheels'' pattern: as the participants' initial predictions improve, the algorithmic advice is useful less often and the frequency in which it is provided decreases.}}
\label{fig:wheels}
\end{figure*}

Figures \ref{fig:wheels-learned} and \ref{fig:wheels-oracle} show the learning effect in the Learned and Omniscient treatments alongside the response of the advising policies to this effect, and  demonstrate the advantage of our two-sided interaction approach: as the human initial prediction improves compared to the algorithmic prediction, the learned advising policy identifies more cases in which the algorithmic advice is not needed, and as a consequence provides significantly less advice.
We note that the better learning observed in the Learned and Omniscient treatments may result from a combination of several effects, which their impact on human learning is not isolated in our experimental design; e.g., the higher informativeness of the given advice and the higher responsiveness to the advice in these treatments. Further studying the factors that facilitate human learning is a broad and interesting direction for future work.

\paragraph{Tension between Imitating the Algorithm and Preserving Complementary Human Strengths.} \label{sec:tension}

The results so far show that participants managed to learn from the algorithmic feedback and improve their initial predictions, and that this improvement was more substantial in the Learned and Omniscient treatments than in the Random and Update treatments. Now we turn to look directly at how the distributions of initial predictions differ between the different treatments. We demonstrate that this learning phenomenon raises a tension between the extent to which humans learn from the algorithmic advice on the one hand, and their ability to preserve their own relative prediction strengths on the other hand.

A comparison of the distribution of human initial predictions and the algorithmic predictions shows that, as expected, in the No Advice treatment, in which human predictions are completely independent of the algorithmic predictions, the distance between these two distributions is the largest (as measured by KL divergence \cite{kullback1997}, see Table \ref{tbl:overview}). The distribution of initial predictions in the Update treatment was the closest to the algorithmic predictions, and the treatments in which the advice was provided only in part of the predictions had intermediate KL divergence values. This suggests that in the Update treatment, in which participants constantly observed the algorithmic risk assessment, the participants learned to predict similar values to the feedback that they observed, while in the partial advice settings this imitation effect was moderated.

A closer look suggests that the partial advice has an advantage in preserving human prediction behavior that is complementary to the predictions of the algorithm. A notable example is that in the always-advising Update treatment participants learned to almost never predict a certain low-risk value of zero -- a value that was never predicted by the algorithm,\footnote{Recall that the algorithm was optimized for minimizing quadratic error, and so avoiding predictions of extreme values is a typical outcome of such an optimization process.} 
but was predicted by human participants in the No Advice treatment in 10.5\% of all predictions. This is despite the fact that in the subset of instances in which humans predicted a zero risk, their predictions were significantly more accurate than their average prediction performance. By contrast, in the Learned treatment participants preserved this relative strength and predicted a risk of zero for 11.0\% of the predictions, and similarly to the No Advice treatment, with a higher accuracy in those predictions relative to their average performance. See more details in Appendix \ref{app:results-responsiveness}. 

\subsection{Fairness}

\begin{figure}
	\centering
		\includegraphics[width=0.6\linewidth]{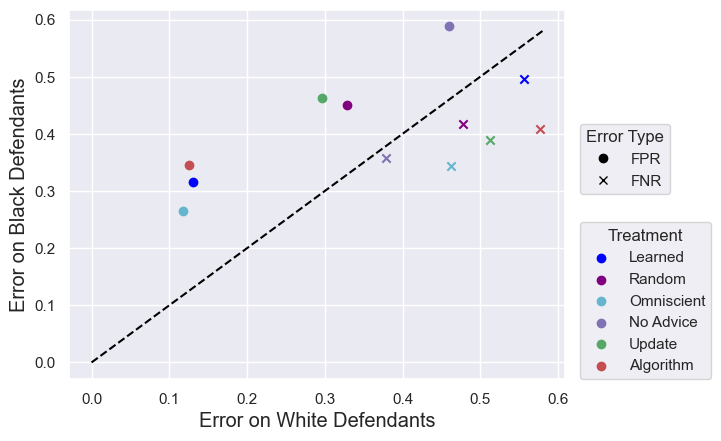}
	\caption{{\small False-positive rates (FPR) and false-negative
rates (FNR) for black and white defendants in each experimental treatment and for the algorithm's predictions. }}
	\label{fig:fpr-fnr}
\end{figure}

We further examine how human decision makers assisted by our advising policies perform with respect to defendants' racial groups. We start by comparing the false-positive rates (FPR) and false-negative rates (FNR) in our experiment for black and white defendants (see Figure \ref{fig:fpr-fnr} and a summary in Table \ref{tbl:overview}, as well as Figure \ref{fig:fairness} in Appendix \ref{app:results-fairness}). 
For each racial group, the group FPR is the rate in which defendants from that group did not violate their release terms but were wrongly classified as high-risk defendants, and the group FNR is the rate in which defendants from the group violated their release terms but were wrongly classified as low-risk defendants. The decision threshold is the value that optimizes F-score \cite{zou2016}, which is 0.3 for each treatment as well as for the algorithm, so that predictions above 0.3 are classified as high-risk decisions and otherwise are classified as low-risk decisions.\footnote{The same threshold is also obtained by taking the fraction of high-risk defendants, which is 0.326 in our dataset (and since risk predictions are in multiples of 0.1).}

First, the results show that the FPR in the Learned treatment, for both black and white defendants, is substantially lower than the FPR in the Update, Random, and No Advice treatments, but at the cost of higher FNR (see Figure \ref{fig:fpr-fnr} as well as   
more details in Appendix \ref{app:results-fairness}). 
Second, in terms of FPR and FNR disparities between black and white defendants, we find that the learned advising policy is comparable to the Update treatment, and has a significant advantage compared with the risk-assessment algorithm. 
See Appendix \ref{app:results-fairness} for the details.

Figure \ref{fig:fpr-fnr} shows that 
according to both FPR and FNR, all treatments have an error that is biased to the same direction which gives harsher predictions for black defendants.
We quantify this discrimination by defining ``classification disparity'' for a treatment as: 
$Pr(Y=0)(FPR_{Black} - FPR_{White}) + Pr(Y=1)(FNR_{White} - FNR_{Black})$. 
The classification disparity 
weighs the discrimination of black compared to white non-risky defendants (the first term), and the bias in favor of white compared to black risky defendants (the second term). Equivalently, the classification disparity can be interpreted in terms of utility from the point of view of the defendant: the first term is the utility gap for non-risky defendants and the second term is the utility gap for risky defendants, both in favor of white defendants and are weighted by the overall frequencies of risky and non-risky defendants in the population.

We find that, interestingly, the algorithm has the highest classification disparity (this is despite the fact that the algorithm did not directly observe the defendant's race whereas the human participants did), the No Advice and Random treatments have the least classification disparity, and Omniscient, Learned, and Update have intermediate disparity levels with an advantage to Omniscient and Learned 
(see Table \ref{tbl:overview}). 
When considering the accuracy-fairness tradeoff, the results show that Learned and Omniscient Pareto dominate the Update treatment (Figure  \ref{fig:tradeoff}  
in Appendix \ref{app:results-fairness}). 
This tradeoff suggests that the use of informed advising policies in Learned and Omniscient allowed the human decision makers on the one hand to extract gains from the high performance of the algorithm, while on the other hand to moderate its racial disparity.  

Finally, in Appendix \ref{app:results-fairness}  
we analyze interaction disparity according to the two measures studied in \cite{benCSCW}, to evaluate whether participants responded to the risk assessment in a racially biased manner. In \cite{benCSCW}, every experimental treatment exhibited disparate interactions, including the Update treatment (which is identical to Update in our experiment) that yielded the smallest disparity. 
Our experiment replicates the results for the first ``influence disparity'' measure for the Update treatment, but this influence disparity was eliminated in the Learned and Omniscient treatments. The second ``deviation disparity'' observed in \cite{benCSCW} was not replicated in our experiment, including in our Update treatment. 

%% file: discussion-ijcai.tex
What is the best way to use algorithms to advise human decision makers? 
Existing methods constantly provide advice and focus on optimizing the algorithmic advice itself or its presentation or explanation to the human user.
Motivated by limitations of the constant-advising approach that frequently advises the users in redundant or even harmful times, and by the complementary abilities of humans and algorithms that have been demonstrated in many settings, 
this paper proposes a responsive advising approach, in which the algorithm interacts with the human user and provides advice only when it is most beneficial for the human in making their decision.

We analyzed over fifty thousand human predictions in five experimental treatments that compared our new responsive-advising approach to the constant-advising approach and to other benchmarks. 
Our analysis shows that people assisted by responsive-advising policies succeeded in making predictions that 
are more accurate 
and better preserve human relative strengths, 
compared with people who constantly received the algorithmic advice. 
Also, human predictions when assisted by our advising policy achieved comparable performance to the algorithm's predictions in terms of accuracy, which, as discussed, is a significant improvement over prior methods,  
and had a significant advantage over the algorithm in terms of fairness measures.
Importantly, we showed that such an advising policy that identifies when (and when not) to provide advice to the human user, based on input from the user, can be automatically learned from existing data.

One basic explanation for the advantage of our approach in terms of accuracy, is that our learned advising policies managed to utilize, for every given prediction problem, both the input from the algorithmic risk assessment and the input from the human user. This resulted in providing the advice in more informative times, and specifically, providing the advice when the algorithmic assessment was more accurate than the human  initial prediction and refraining from providing misleading advice. 
Notably, in our implementation, the input from the user was composed of solely the human's initial prediction. The results show that this single additional bit of information that the AI system received already enabled this significant advantage. Future work will determine whether more complex inputs from the human users can further improve the quality of the advising policies and their usefulness for the users (e.g., by using active queries to the users, individualized analyses of their historical behavior, or signals that indicate their levels of confidence or engagement).

Aside from the direct impact on performance, our analysis raises two concerns about the longer-term impact on human decisions from constantly receiving input from an algorithm. 
First, in the treatment where humans received algorithmic advice all the time, this advice was followed by a weak response, and importantly humans often failed to identify those cases where this advice was especially useful for them. 
By contrast, in the treatments where advice was given only in selected times, this advice was followed by higher responsiveness on the side of the human decision makers. 
Indeed, in a within-treatment analysis in the Random treatment, we find a clear connection between the frequency in which the advice is provided and human responsiveness to the advice, which we term the ``scarcity effect'':  When advice is given less frequently, it tends to be followed by stronger responses. 
We conjecture that observing  advice more frequently leads to habituation in human responses, while scarce advice are perceived as more valuable, however further research is needed in order to explain the behavioral source of this effect. 
More broadly, our study suggests the importance of studying the effect of partial or conditional advising on advice utilization, which in contrast to various other factors (see, e.g., review in \cite{bonaccio2006}) has not yet received attention in behavioral literature.

Second, our analysis of the initial predictions given by the participants in the experiment (i.e., before observing the algorithmic risk assessment), shows that people who constantly observe the algorithmic advice learn to imitate the algorithm's past predictions.  Arguably, this is instead of focusing on forming their own judgments for the problems at hand. 
By contrast, in the interactive advising treatments, in which advice was provided in only about one third of the times,  
this imitation effect was moderated, and the distributions of human-predicted risk assessments preserved features that were unique to human judgments (and not to the algorithm), which almost completely disappeared in the constant-advising treatment. 
This empirical observation of the imitation effect raises a concern, which may in fact be inherent to any algorithmic advising setting: on the one hand algorithmic advice assists humans to improve their decisions, but on the other hand, through repeated exposure to the algorithm, human decision makers may also internalize its biases and weaknesses into their own judgments.
Our results suggest that the advising-policy approach that we propose manages to balance this tradeoff to a good extent.

A potential 
limitation of the present study is that the findings are based on predictions made by Mechanical Turk workers in controlled experimental settings, rather than on decisions made in practice by real human experts like judges or clinicians. While controlled experiments with lay decision makers are useful in isolating and suggesting human behavioral tendencies that are then identified in practice \cite{guthrie2000,barberis2013}, the effects in the ``real world'' may differ from those in experimental settings due to the experimental abstracted context and the decision makers' domain knowledge and levels of expertise \cite{tschandl2020,zhang2020}. Thus, continued research is important in order to study the extent to which the findings generalize to human-algorithm interactions in practice, and specifically to test the usefulness of our responsive algorithmic-advising approach in real AI-assisted decision making scenarios.

Algorithmic advising systems are becoming increasingly prevalent in situations in which human judgment is important and cannot be replaced by an algorithm. Such systems provide advice to human decision makers in high-stake domains 
ranging from criminal justice to finance and healthcare,  
as well as in day-to-day applications such as personal assistants and recommendation systems. 
The ways in which we choose to design such algorithmic advising tools shape our lives and may have broad implications to society. 
The present study points to the importance of asking {\em when} algorithms should provide advice. 
The findings show that 
a responsive approach that considers input from the human user and provides advice that is 
\emph{focused} on those places where it is most needed 
can better assist humans in making their decisions. 
Future work will study how to best apply this approach in current AI-assisted decision making systems,  
aiming to create better human-AI collaboration that will efficiently harness AI strengths to assist humans in making better decisions.

%% file: ethical_statement.tex

This study deals with the use of artificial intelligence (AI) to aid human decision making. It was reviewed and approved by the Harvard University Institutional Review Board (IRB) (under the reference number IRB21-0851) and the National Archive of Criminal Justice Data (NACJD)(data usage agreement DUA21-0771) to ensure that ethical considerations were addressed during the research process.

AI-assisted decision making has the potential to bring many benefits, but it also poses ethical risks. One concern that may arise is about the potential misuse of the technology to bias people to make decisions in directions that are not beneficial for themselves. Our approach attempts to mitigate this risk at the algorithmic level by decoupling the advising policy from the risk-assessment algorithm, instead of training a single end-to-end algorithmic component. This separation constrains the risk-assessment algorithm to be trained to optimize the quality of its risk assessments independently of the human decision maker, rather than learning to provide biased assessments that aim to affect their decisions.  Our advising policy aligns well with the decision maker's objective, providing advice only when it deems the algorithmic assessment is more accurate than the human prediction. Another potential risk is that the AI assistance may unintentionally lead to unfair outcomes and contribute to discrimination against certain groups of people. 
We analyzed our results for such potential biases with respect to defendants' racial groups. The analysis suggests that the informed advising policies manage to balance the impact of algorithmic advice on human predictions by gaining from the algorithm's high performance while reducing its racial disparity. 

The use of AI technology in decision making more broadly, may raise concerns about removing human agency and responsibility from the decision-making process. When algorithms alone are making the decisions, it may be difficult to hold anyone accountable for the outcomes. Moreover, lack of transparency in how these algorithms work can make it difficult to understand the rationale behind specific decisions. In this study, we focus on the setting in which the algorithm only provides advice, but the human makes the final decision. Furthermore, we use a decision structure that encourages humans to form an independent opinion: the human first makes a prediction on his own and only then observes the algorithmic advice. However, our present study does not address the issue of transparency, except for a high-level explanation of the machine learning algorithm used. Explanatory information from the algorithm regarding its decisions in combination with our advising-policy approach is an interesting extension for future studies.

%% file: related-work.tex
Collaboration between AI and humans in decision making has been studied in the machine learning literature in a framework called ``human-in-the-loop'' machine learning. This framework considers settings in which an algorithmic system utilizes feedback from humans in order to improve its performance on a given problem, including also settings of crowd-supported machine learning  \cite{kamar2012,cheng2015flock,wang2019crowd} and rejection-learning settings in which algorithms make sequences of decisions and have the alternative to defer some of their decisions to human decision makers  \cite{madras2018predict,mozannar2020consistent,wilder2020learning} (for further background on rejection learning, see \cite{bartlett2008classification,cortes2016learning,hendrickx2021machine}). 
In contrast to these settings, in the setting that we study the final decisions are made by human decision makers and the role of the AI is to assist the human to improve these decisions.

Another line of research that deals with human-AI interactions studies explainable or interpretable machine learning in a broad range of scenarios, focusing mainly on complex learning models such as deep-learning \cite{samek2017explainable,singh2020explainable} and  reinforcement learning \cite{finkelstein2022explainable,cohen2022finding,heuillet2021explainability,septon2022integrating} (for further background, see \cite{arrieta2020explainable}).   
However, as discussed in the introduction, in human decision-making settings it has been shown empirically  that interpretable AI does not necessarily improve the ability of humans to beneficially  utilize AI recommendations or  monitor their quality \cite{poursabzi2021,bansal2021does}. 
In this context, it could be that our model in which algorithmic advice is presented only for those problems in which it is most useful may lead human decision makers to perceive the AI advice as simpler to interpret and understand. However, we do not study interpretability directly and further work is required to determine if this  indeed is the case.

More broadly, different types of human-AI interactions that have received recent attention deal with strategic settings. These include algorithmic predictions of human decisions in strategic interactions \cite{kolumbus2019neural,noti2016behavior,noti2021bid,hartford2016deep} and, in a different line of work, models of strategic relations between humans and AI algorithms. The latter models describe settings in which AI systems make decisions that impact the welfare of humans, and the humans, typically modeled in this line of literature as rational game-theoretic agents, manipulate the data that they provide to the AI \cite{cai2015optimum,hardt2016strategic,chen2019learning,dong2018strategic,ghalme2021strategic} or strategically control the AI agent's parameters \cite{kolumbus2022auctions,kolumbus2021and} to promote their own interests. 
In contrast to these strategic settings, in our setting, the goals of the human decision-maker and the AI system are essentially aligned. Both the AI and the human jointly aim to perform the same task. The challenge in our case, therefore, lies not in balancing diverging strategic interests but rather in optimizing the interaction to allow the human to extract the most from their collaboration with the AI and improve their own predictions.

%% file: setting.tex
We demonstrate our approach in the pretrial-release decision setting.
When a person in the United States is arrested and brought in front of a judge, the judge has to decide whether to release or detain the defendant until his trial. This decision is based on the judge's assessment of how likely is the defendant, if released, to violate his release terms, i.e., to commit another crime until his trial or to fail to appear in court for his trial. 
In the United States, judges are assisted by risk-assessment algorithms for making this important decision \cite{courts2020,pji2019,compas2019}, and prior works have studied the algorithms' prediction performance \cite{kleinberg2017,angwin2016} as well as the impact of using algorithmic risk assessment on human decision  makers -- both with judges and with nonexperts in experimental settings \cite{benFAT,benCSCW,albright2019}.

We use a dataset collected by the U.S. Department of Justice, that contains records of 151,461 criminal defendants who were arrested between the years 1990 and 2009, in 40 of the 75 most populous counties in the United States \cite{data2014}. This dataset was also used in previous studies (e.g., in \cite{benFAT,benCSCW,marx2020,liu2021}). The data include information about arrest charges, demographic characteristics, criminal history, pretrial release and detention, adjudication, and sentencing. We use the cleaned version of the dataset generated and used by \cite{benFAT,benCSCW}, that contains records of 47,141 defendants that remained after restricting the analysis to defendants who were released before trial, who were at least 18 years old, and whose race was recorded as either black or white. Thus, the dataset we use includes only defendants who were released before trial and we thus know the ground truth information about their pretrial-release outcome, i.e., we know for each case whether eventually the defendant was violating his release terms. Among the defendants in this dataset, 29.8\% violated their pretrial-release terms.

%% file: app-results.tex
As described in the main text, we learned the advising policy by using data from the experiment of \cite{benFAT}. The data consist of 6,250 predictions made by 250 human participants on Mechanical Turk, for 500 defendant cases, in the control treatment in which the participants did not observe the algorithmic risk assessment. To better adapt to our target domain, which is a new experiment in which humans interact with a learned advising policy rather than predict independently from it, we trained our model on an augmented version of this dataset. We augmented the dataset in two steps: First, we added simulated defendant cases, such that each original defendant record was duplicated six times on all attributes except for the age attribute that was set to $\pm3$ years from the original age value (but still restricting the age to be above 18). This step augmented the dataset from 6250 predictions for 500 cases to 40,529 predictions for 3251 cases. This augmentation is based on the assumption, which we also verified in the data, that a slight variation in a defendant age would barely affect the prediction. Second, for each defendant case in the augmented set of cases, we simulated human predictions by sampling from a smoothed version of the empirical distribution of human predictions on this case. This step doubled the size of the dataset, from 40,529 to 81,058 predictions.  

We trained a random forest binary classifier on these augmented data by using the {\em scikit-learn} python package. An optimization process of the model's hyperparameters to obtain the best accuracy  in a 20-fold cross-validation on the training data  gave the following hyperparameters: 
$\texttt{n\_estimators}=400$, $\texttt{min\_samples\_splits}=100$, 
$\texttt{min\_samples\_leaf}=50$, and $\texttt{max\_features}=4$.
We also fitted the model's threshold, and used a value of $0.42$, which calibrates the advice frequency on the training data. That is, by using a model's threshold of $0.42$, the model's advice frequency on the training data matched the 33.3\% advice frequency according to the ground truth.

\section{Additional Details on the Experimental Setup} \label{app:setup}

We recruited participants on Amazon Mechanical Turk, restricting the participation to Mechanical Turk workers inside the United States who had an historical acceptance rate of at least 75\%. All Mechanical Turk workers are at least 18 years old. 

The experimental procedure is similar to that of \cite{benCSCW}. Upon arrival, participants read a brief description of ``what to expect'' in the experiment, and were asked to sign a consent form. Then, they were randomly assigned to one of the five experimental treatments: Learned, Omniscient, Random, No Advice, or Update. The experiment started with a tutorial, followed by a short intro survey, the primary prediction task of predicting risk for a series of 50 criminal defendants, and an exit survey to obtain participants' reflection on the task. The series of 50 defendants was drawn at random for each participant, from the same sample of 300 defendant cases that was used in the experiment of \cite{benCSCW}. The tutorial was visible at the bottom of the screen throughout the entire prediction task so that participants could look up background information and the definitions of key terms.

In the primary prediction task, participants were presented with descriptions of the 50 defendants, one by one. 
Each description included seven features: age, gender, race, offense type, number of prior arrests, number of prior convictions, and previous failure to appear. 
For example: ``{\em Defendant \#1 is a 35 year old black male. He was arrested for a property crime. The defendant has previously been arrested 10 times. The defendant has previously been released before trial, and has never failed to appear. He has previously been convicted 3 times.}'' Then, they were asked to predict the defendant's risk on a scale from 0\% to 100\%, in intervals of 10\%: ``{\em How likely is this defendant to fail to appear in court for trial or get arrested before trial?}''

In the three algorithmic-assistant treatments---Learned, Omniscient, and Random---after making each prediction, the algorithmic assistant decided whether to advise the participant with the algorithmic risk assessment, as follows. 
In Learned, our learned advising policy determined whether or not to advise based on its prediction of how likely the algorithmic risk assessment is to be more accurate than the participant's prediction. 
In Omniscient, an advising policy that uses hindsight information on the defendant's true pretrial-release outcome decided to show the advice exactly in those cases where the algorithmic risk assessment was more accurate than the participant's prediction. 
In Random, an advising policy decided to show the advice with probability of 30\%, which corresponds to the advice frequency of our learned advising policy on the training data, and did not provide advice when the algorithmic risk assessment and the participant's prediction were identical. The instructions informed the participants that the set of cases in which the algorithmic assistant provides the advice is randomized.

When the decision was not to advise, the participant continued directly to the next defendant case. When the decision was to advise, the participant was presented with a message such as: ``{\em Your algorithmic assistant identifies that your current prediction is likely to have high error, and advises you to improve the prediction to 40\%.\footnote{As in \cite{benCSCW}, because the participants predicted risk in increments of 10\%, we rounded the algorithmic risk predictions to the nearest 10\% when presenting them to participants and in the analysis of the results.}}'' In Random, the message was: ``{\em Your algorithmic assistant predicts that this person is 40\% likely to fail to appear in court for trial or get arrested before trial.}''
Then, the participant was asked to make his final prediction, by either accepting the algorithmic advice or editing his prediction. In the No Advice treatment there was no additional information regarding the risk assessment, and after making each prediction the participant  continued directly to the next defendant case. The Update treatment was identical to that of \cite{benCSCW}: after each prediction that the participant made, he was shown the algorithm's risk prediction and was free to update his prediction or to stay with his original prediction.

\begin{table*}
\centering
\caption{Demographic and general details of the participants in the experiment.}
\begin{tabular}{lrrrrr}
  & Learned & Random & Omniscient & No Advice & Update \\
	& N=218 & N=200 & N=200 & N=258 & N=220 \\
\midrule


Male & 64.2\% & 61.0\% & 64.5\% & 64.3\% & 65.5\% \\
Black & 12.4\% & 7.5\% & 16.0\% & 10.9\% & 10.0\% \\
White & 81.2\% & 77.0\% & 76.5\% & 76.4\% & 80.0\% \\

18-24 years old & 4.1\% & 4.5\% & 3.5\% & 3.1\% & 1.4\% \\
25-34 years old & 41.3\% & 42.0\% & 42.0\% & 42.2\% & 46.8\% \\
35-59 years old & 50.0\% & 48.5\% & 49.0\% & 48.4\% & 49.1\% \\
60+ years old & 4.6\% & 5.0\% & 5.5\% & 6.2\% & 2.7\% \\

College degree or higher & 81.7\% & 87.0\% & 85.0\% & 90.3\% & 83.6\% \\
Criminal justice familiarity (1-5 scale) & 3.3 & 3.2 & 3.5 & 3.3 & 3.2 \\
Machine learning familiarity (1-5 scale) & 3.3 & 3.1 & 3.4 & 3.2 & 3.1 \\

Participant payment & \$4.35 & \$4.27 & \$4.45 & \$4.17 & \$4.30 \\
Experiment duration (minutes) & 19.9 & 22.3 & 23.2 & 20.1 & 23.3 \\
Experiment clarity (1-5 scale) & 4.33 & 4.28 & 4.25 & 4.41 & 4.39 \\
Experiment enjoyment (1-5 scale) & 3.83 & 3.85 & 3.82 & 3.90 & 3.87 \\

\bottomrule
\end{tabular}
\label{tbl:demographics}
\end{table*}

Upon completion of the experimental trial, the participants received a base payment of \$2 plus a performance-dependent bonus of up to \$3. The average duration of an experimental trial was 21.6 minutes. The bonus was computed additively over the predictions. The reward for each prediction was determined according to a Brier score function: $score=1-(prediction-outcome)^2$, normalized such that a perfect predictive accuracy on all 50 predictions would yield a total bonus of \$3. The average bonus in the experiment is \$2.31. The Brier score is a proper score function \cite{gneiting2007}, and thus 
it incentivizes the users 
to report their true risk estimates. In the tutorial we explained this truthfulness property, and also included a comprehension question about it to verify  understanding.

The intro and exit surveys included a simple attention question, and we excluded from our analysis participants who failed to answer correctly both attention questions. Also, we included in our analysis only participants who completed the prediction task for the full series of 50 defendant cases, and allowed a single participation for each worker.  
In total, 1,096 workers are included in the analysis. See Table \ref{tbl:demographics} for more details on the participants in each treatment.

\section{Results: Additional Details} \label{app:results}


\begin{figure*}[!t]
\centering
\begin{subfigure}{.19\linewidth}
\center
\includegraphics[width=1.0\linewidth]{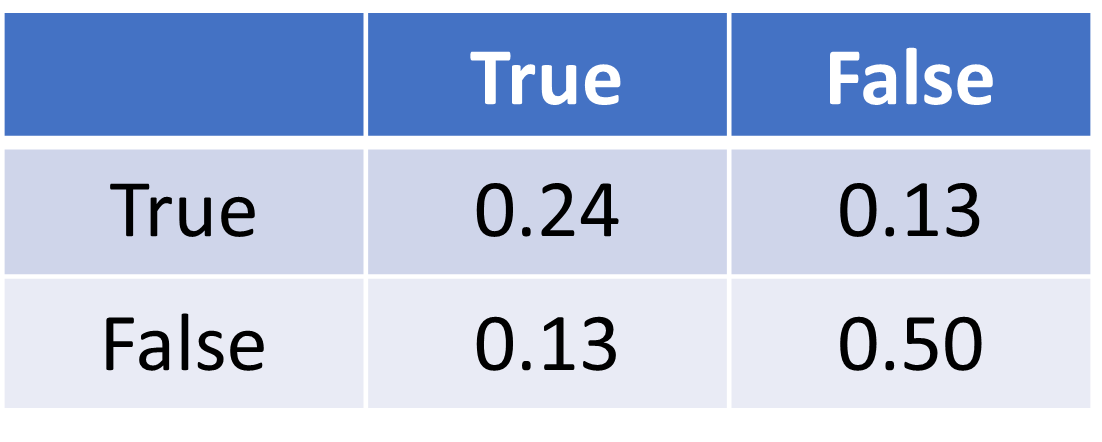}
\caption{Learned}
\label{fig:confusion-matrices-learned}
\end{subfigure}
\begin{subfigure}{.19\linewidth}
\center
\includegraphics[width=1.0\linewidth]{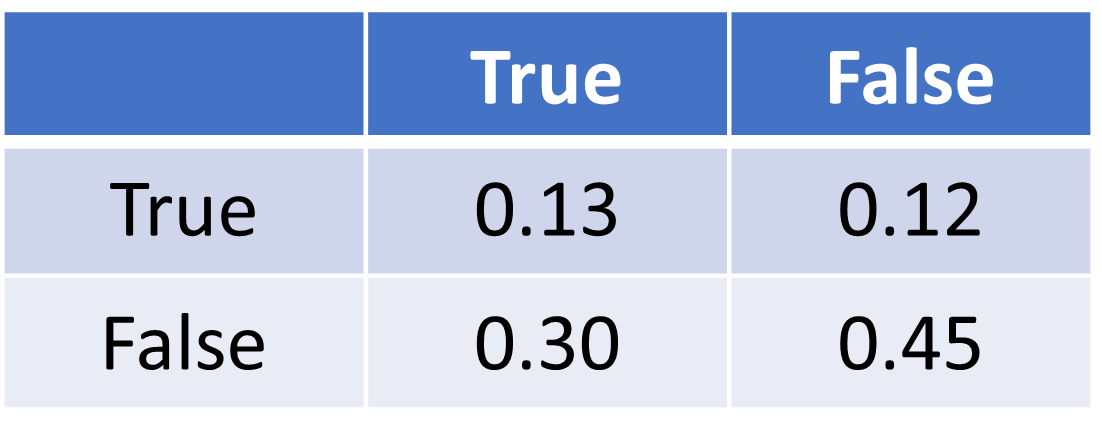}
\caption{Random}
\label{fig:confusion-matrices-random}
\end{subfigure}
\begin{subfigure}{.19\linewidth}
\center
\includegraphics[width=1.0\linewidth]{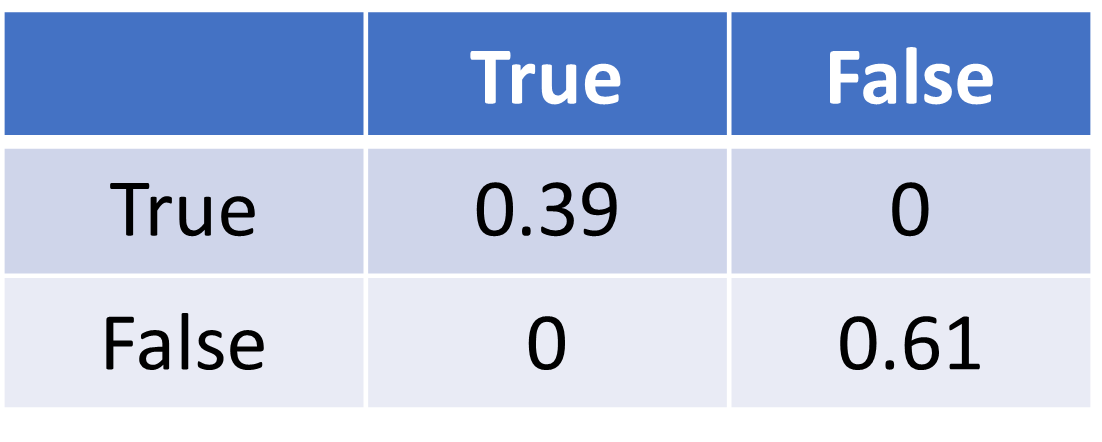}
\caption{Omniscient}
\label{fig:confusion-matrices-oracle}
\end{subfigure}
\begin{subfigure}{.19\linewidth}
\center
\includegraphics[width=1.0\linewidth]{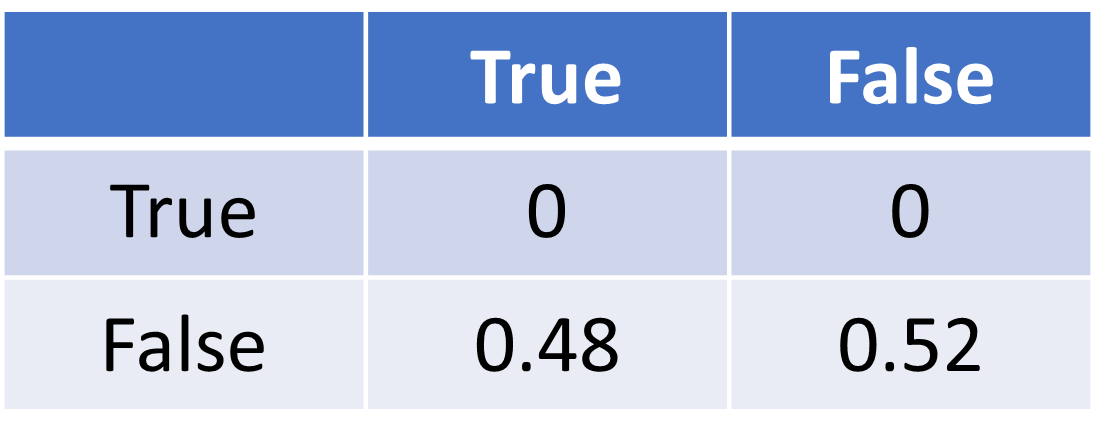}
\caption{No Advice}
\label{fig:confusion-matrices-baseline}
\end{subfigure}
\begin{subfigure}{.19\linewidth}
\center
\includegraphics[width=1.0\linewidth]{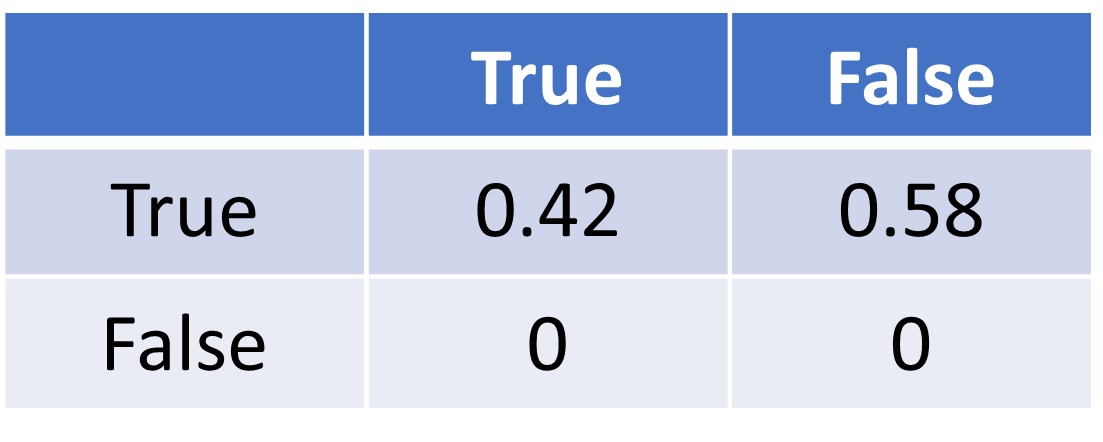}
\caption{Update}
\label{fig:confusion-matrices-update}
\end{subfigure}
\caption{{\small Confusion matrices. For each treatment, the left and right columns show the ground-truth rates of whether or not the algorithmic risk prediction was more accurate than the human initial prediction,   
respectively, and the top and bottom rows show the rates in which the advising policy determined to provide or not to provide the advice, respectively.}}
\label{fig:confusion-matrices}
\end{figure*}	

\begin{figure*}[!t]
	\centering
		\includegraphics[width=1.0\linewidth]{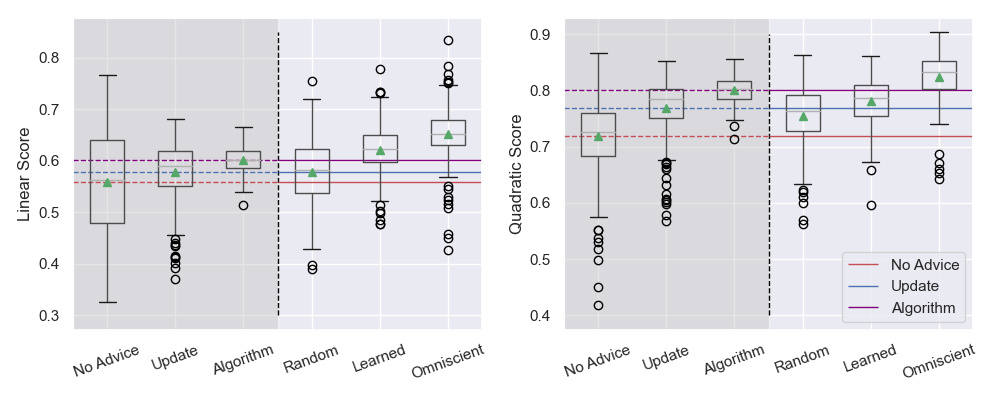}
		\caption{Distributions of participant performance in the different experimental treatments, and of the algorithm's performance, according to both linear and quadratic scores.  The performance is presented via standard box plots (Tukey style \protect\cite{tukey1978}), where the box extends from the first quartile to the third quartile of the data, the gray line indicates the median and the green triangle indicates the mean. The No Advice, Update, and algorithmic benchmarks are presented on the left of each figure, and for ease of comparison, their means continue as horizontal lines. The algorithm's performance is computed over its predictions for the cases given to participants in the Learned treatment.}
	\label{fig:score-dists}
\end{figure*}

\subsection{Overview}
Table \ref{tbl:demographics} presents demographic and general details and Table 1 in the main text 
presents an overview of the results of the main metrics in the five experimental treatments.


\subsection{Learning Performance}
Figure \ref{fig:confusion-matrices} shows the confusion matrices in the different experimental treatments. By summing the left column for each treatment, we obtain the empirical frequency in which the risk assessment was more accurate than the initial predictions made by human participants in the treatment. Interestingly, it can be seen that this rate is significantly lower in the Learned and Omniscient treatments than in all other treatments. This shows differences in behavior of the participants in these treatments when they gave their initial predictions, i.e., before they observed any algorithmic input (and specifically, an improvement in their prediction relative to the algorithm), which is an indication of human learning. Later in our analysis (in the main text and in the present appendix) 
we further analyze this difference in behavior, and show that the use of informed advising policies facilitates human learning.



\subsection{Impact on Human Prediction Performance} \label{app:results-impact}
\begin{figure*}[!t]
\centering
\begin{subfigure}{0.49\linewidth}
\center
\includegraphics[width=1.0\linewidth]{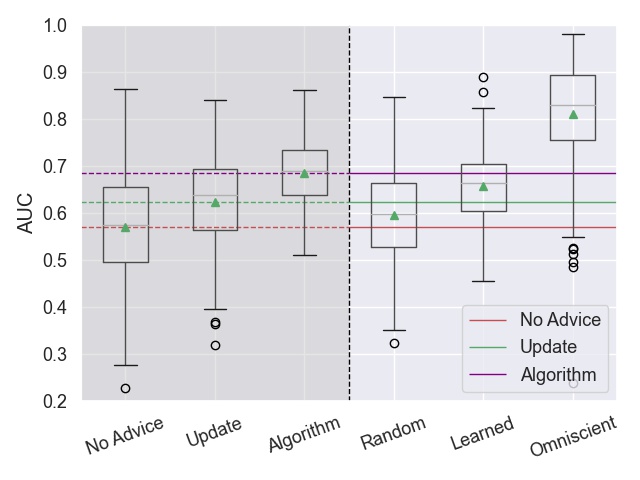}
\caption{{\small Performance distributions.}}
\end{subfigure}
\begin{subfigure}{0.49\linewidth}
\center
\includegraphics[width=1.0\linewidth]{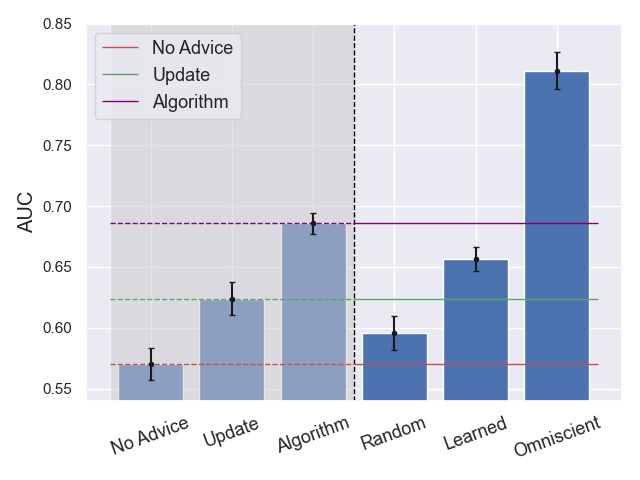}
\caption{{\small 95\% confidence intervals.}}
\end{subfigure}
\caption{Performance according to AUC.}
\label{fig:auc}
\end{figure*}

\begin{figure*}[!ht]
\centering
\begin{subfigure}{0.49\linewidth}
\center
\includegraphics[width=1.0\linewidth]{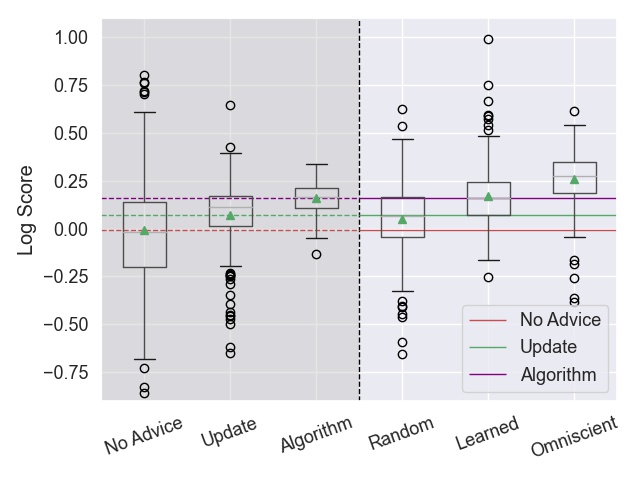}
\caption{{\small Performance distributions.}}
\end{subfigure}
\begin{subfigure}{0.49\linewidth}
\center
\includegraphics[width=1.0\linewidth]{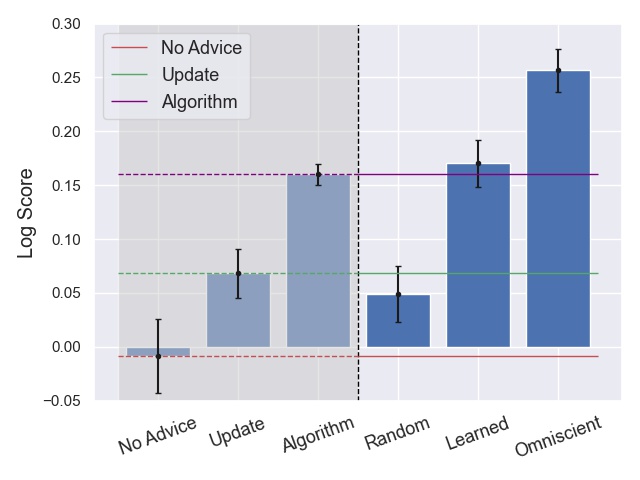}
\caption{{\small 95\% confidence intervals.}}
\end{subfigure}
\caption{Performance according to the logarithmic score.}
\label{fig:logscore}
\end{figure*}

In the main text we evaluate the prediction success of the human decision makers in our experiment according to the linear score and the quadratic score. Figure \ref{fig:score-dists} presents the distributions of the average scores of the participants in the different experimental treatments and of the algorithmic predictions. In addition, Figures \ref{fig:auc} and \ref{fig:logscore} present the performance according to other measures: the AUC and the logarithmic score, respectively.
For each measure, the figures present the full distributions of participant and algorithmic performance, via standard box plots as in Figure \ref{fig:score-dists}, as well as bar charts with error bars that indicate 95\% confidence intervals.

\subsection{Interaction between the Human Decision Maker and the Algorithm} \label{app:results-responsiveness}

\begin{figure*}[!t]
\centering
\begin{subfigure}{0.49\linewidth}
\center
\includegraphics[width=1.0\linewidth]{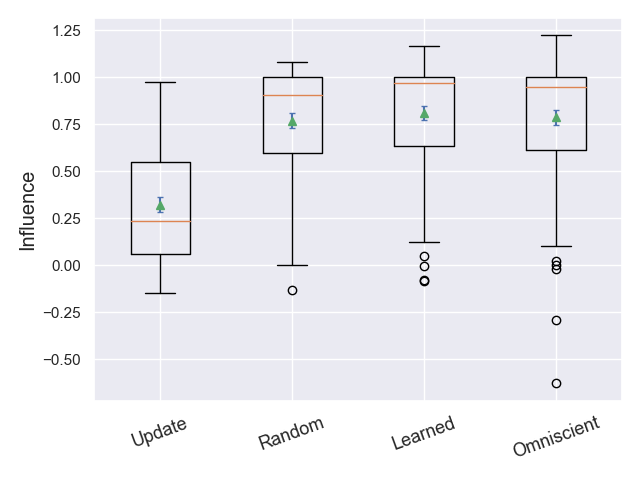}
\caption{{\small }}
\label{fig:influence}
\end{subfigure}
\begin{subfigure}{0.49\linewidth}
\center
\includegraphics[width=1.0\linewidth]{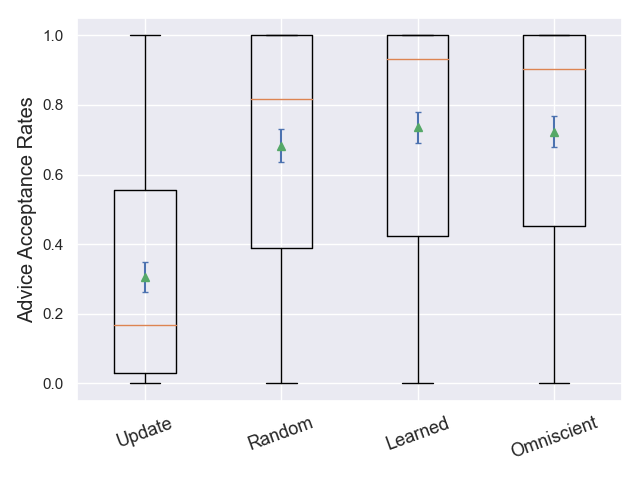}
\caption{{\small }}
\label{fig:acceptance-rates}
\end{subfigure}
\caption{Human responsiveness to the algorithmic advice.}
\label{fig:responsiveness}
\end{figure*}

\paragraph{Human Responsiveness to the Algorithmic Advice:}
Figures \ref{fig:influence} and \ref{fig:acceptance-rates} show the distributions of the advice influence and the advice acceptance rate measures, respectively, over participants, in each of the four treatments in which the advice is given. As can be seen, according to both measures, the responsiveness to the advice is much stronger
in all the three algorithmic-assistant treatments, in which the advice was provided only for part of the predictions (i.e., Random, Learned, and Omniscient), compared with the Update treatment in which the algorithmic risk assessment was provided for all predictions. Although the responsiveness to the advice is lower in Random than in Learned and Omniscient, the differences between these three treatments are not statistically significant.

\FloatBarrier

\FloatBarrier

The large gap between Update and the three experimental treatments could result from two potential sources: the partial advice and a potential framing  
in the algorithmic-assistant setting. 
To   
isolate the effect of the partial advising, we look within the Random treatment. As mentioned in the main text, in the Random treatment  (unlike the Learned and Omniscient treatments) the variance in the frequencies in which participants received the advice comes from getting the advice with a fixed probability in each prediction.    
Figure 3 in the main text  
shows a significant pattern, according to both responsiveness measures, which we term a ``scarcity effect'': human responsiveness to the advice tends to increase with the scarcity of the advice. Specifically, the advice frequency is negatively correlated with the responsiveness of participants to the advice, as measured by the advice acceptance rate ($\rho=-0.23$, $p < 0.001$) and by the advice influence measure ($\rho=-0.29$, $p < 0.0001$).

\paragraph{Indication of Human Learning:}

\begin{figure*}[!t]
\centering
\begin{subfigure}{0.49\linewidth}
\center
\includegraphics[width=1.0\linewidth]{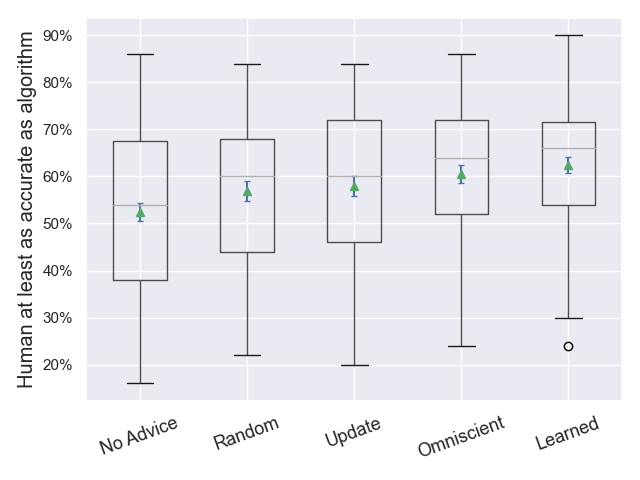}
\caption{{\small Full experiment.}}
\label{fig:learning-overall}
\end{subfigure}
\begin{subfigure}{0.49\linewidth}
\center
\includegraphics[width=1.0\linewidth]{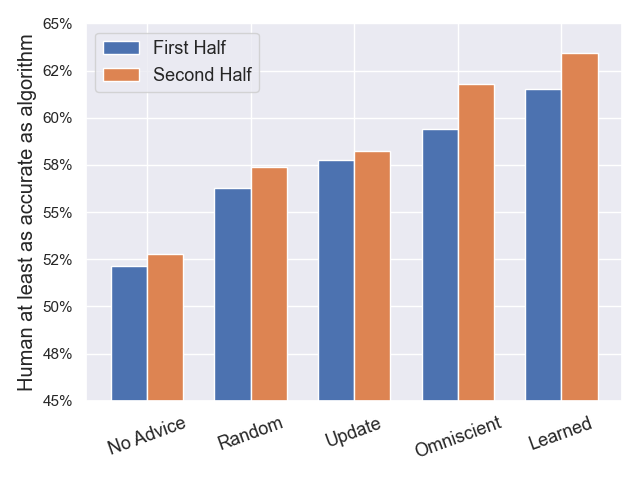}
\caption{{\small First vs. second halves of the experiment.}}
\label{fig:learning-over-time}
\end{subfigure}
\caption{Indication of human learning. The y-axis is the frequency of participants' initial predictions that were at least as accurate as the algorithmic prediction. (\ref{fig:learning-overall}) Distributions over participants in each treatment. (\ref{fig:learning-over-time}) Average frequency in first and second halves of the experiment in each treatment. Differences between first and second halves are statistically significant only for Learned and Omniscient (paired t-test, $p<0.04$ and $p<0.01$, respectively).}
\label{fig:human-learning}
\end{figure*}

Figure \ref{fig:human-learning} shows the performance of participants' initial (i.e., unassisted) predictions in terms of the frequency of initial predictions that were at least as accurate as the algorithmic prediction. 
Figure \ref{fig:learning-overall} presents the performance distributions over participants in each treatment. In all experimental treatments this frequency is significantly higher than in the No Advice treatment ($p<0.01$). 
In the Learned and Omniscient treatments, human initial predictions were at least as accurate as the algorithmic risk score in 62.5\% and 60.6\% of the predictions, respectively, which is significantly higher than in all other treatments: 58.0\%, 56.8\%, and 52.5\%, in Update, Random, and No Advice treatments, respectively ($p<0.02$, except for the comparison of Omniscient and Update for which $p=0.072$). The advantage of Update over the No Advice benchmark is consistent with the indication of learning in Update compared to No Advice that was observed in \cite{benCSCW}.

We find significant learning over time only in the Learned and Omniscient treatments. Specifically, we compared changes over time in the frequency in which the human initial prediction was at least as accurate as the algorithm's risk score.	We found that only in the Learned and Omniscient treatments this frequency significantly improved between the first and second halves of the series of 50 predictions that participants made (paired t-test, $p<0.04$, Figure \ref{fig:learning-over-time}). In addition, the average frequency was positively correlated with the prediction period only in the Learned and Omniscient treatments ($\rho=0.35$ $p<0.02$, and $\rho=0.49$ $p<0.001$, respectively).


\begin{figure*}[!t]
\centering
\begin{subfigure}{0.49\linewidth}
\center
\includegraphics[width=1.0\linewidth]{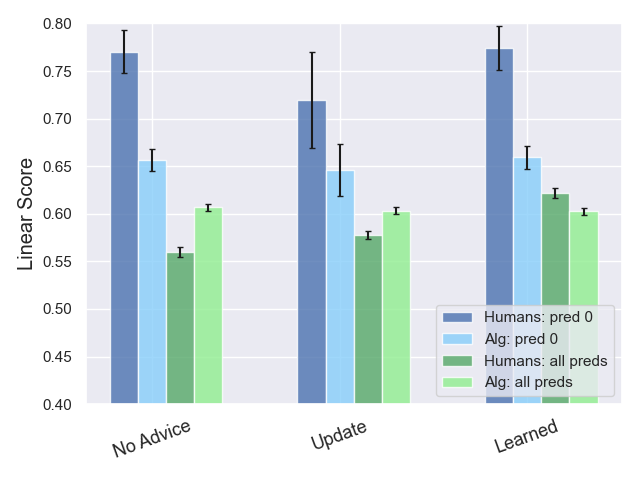}
\caption{{\small}}
\label{fig:pred0-linear}
\end{subfigure}
\begin{subfigure}{0.49\linewidth}
\center
\includegraphics[width=1.0\linewidth]{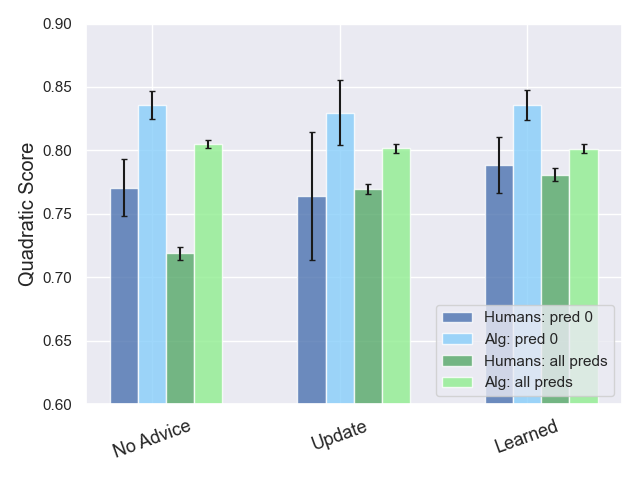}
\caption{{\small}}
\label{fig:pred0-quadratic}
\end{subfigure}
\caption{A comparison of human and algorithmic performance according to the linear and quadratic scores, in the cases in which the human initial prediction is zero and over all predictions, in Learned, No Advice, and Update (the performance is according to the final prediction the human made).}
\label{fig:pred0}
\end{figure*}

Finally, we find that the learning effect we observed was only weakly translated to an improvement in the quality of the initial predictions with respect to the ground truth. That is, testing over all predictions, only in the Learned and Omniscient treatments the initial predictions participants made outperformed those in the No Advice treatments according to both the linear and quadratic scores ($p<0.01$), and testing for improvement over time we find that only the Omniscient treatment consistently improves such that both score measures were positively correlated with prediction period ($\rho=0.33$ $p<0.03$, and $\rho=0.45$ $p<0.01$, for linear and quadratic scores, respectively). 
This weaker effect in the comparison with the ground truth could be expected due to multiple noise sources in the learning process (humans imperfectly learn from imperfect algorithmic feedback about the ground truth),  
and it may also be that the time horizon is not long enough for the learning effect to become evident.

\begin{figure*}[!t]
\centering
\begin{subfigure}{.32\linewidth}
\center
\includegraphics[width=1.0\linewidth]{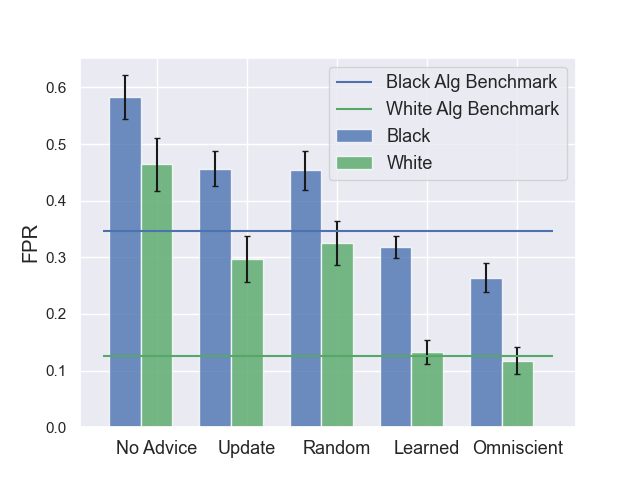}
\caption{}
\label{fig:fpr}
\end{subfigure}
\begin{subfigure}{.32\linewidth}
\includegraphics[width=1.0\linewidth]{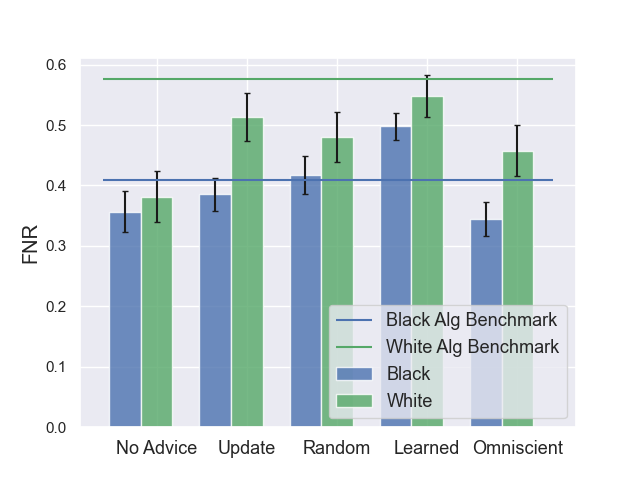}
\caption{}  
\label{fig:fnr}
\end{subfigure}
\begin{subfigure}{.32\linewidth}
\center
\includegraphics[width=1.0\linewidth]{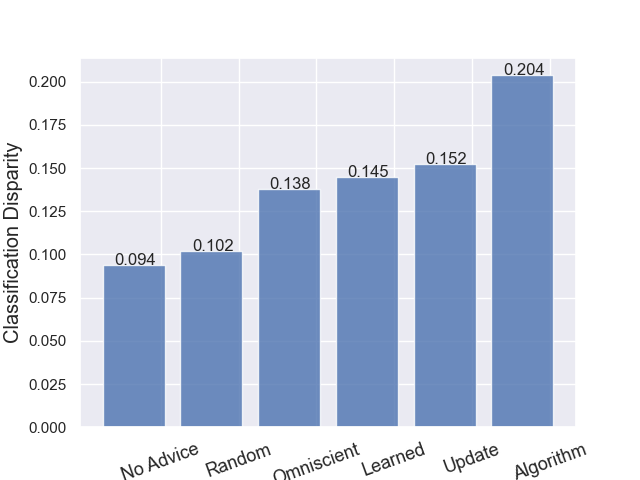}
\caption{}
\label{fig:disparity}
\end{subfigure}
\caption{{\small Performance with respect to defendants' racial groups. (a) False-positive rates (FPR); (b) False-negative rates (FNR); (c) Classification disparity. }}
\label{fig:fairness}
\end{figure*}


\paragraph{Tension between Imitating the Algorithm and Preserving Complementary Human Strengths:}

In the main text we focus on one behavior that we observed -- predicting a risk of zero -- that is very different between the humans and the algorithm. The algorithm never predicts a risk of zero, while in No Advice a risk of zero was predicted by human participants in 10.5\% of all predictions. This behavior remains in Learned (11\% of all initial predictions are zero), but in Update people learn not to predict this value (only 2.17\% of all predictions are zero). 

Figures \ref{fig:pred0-linear} and \ref{fig:pred0-quadratic} compare human and algorithmic performance according to the linear and quadratic scores, respectively, in the cases in which the human initial prediction is zero and over all predictions, in Learned, No Advice, and Update (the performance is according to the final prediction the human made). 
As can be seen, in No Advice and Learned when people predict zero they tend to do better than their overall performance and thus it seems that learning not to predict zero in Update hurts their performance. Also, the comparison of the algorithm with No Advice shows that when human initial prediction is zero (unlike when considering all predictions) it is not clear whether the algorithm is at all better than the human, and the question which behavior is more desirable depends on the measure we wish to optimize.


\subsection{Fairness} \label{app:results-fairness} 

\begin{figure*}
	\centering
		\includegraphics[width=0.8\linewidth]{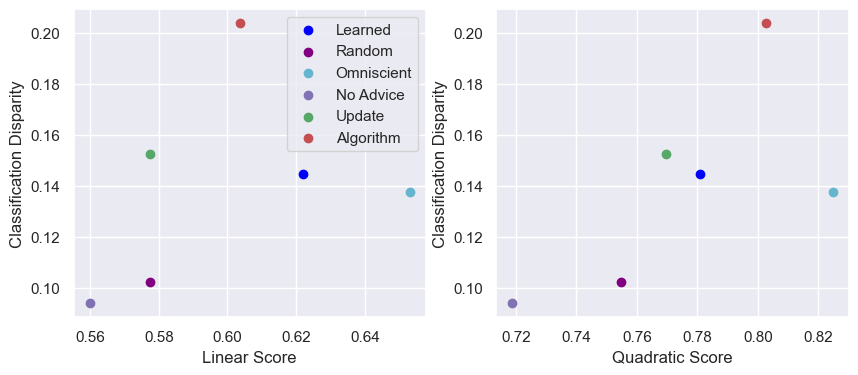}
	\caption{{\small Accuracy-fairness tradeoff. }}
	\label{fig:tradeoff}
\end{figure*}

Figures \ref{fig:fpr} and \ref{fig:fnr} show, for black and white defendants, the group false-positive rates (FPR) and group false-negative rates (FNR) of the predictions made by participants in the different experimental treatments, as well as those obtained from the algorithm. 
See also Figure 5 
and a summary in Table 
1 in the main text.

We start by evaluating the levels of group FPR (Figure \ref{fig:fpr}) and group FNR (Figure \ref{fig:fnr}) in the Learned treatment. The results show that the FPR in the Learned treatment, for both black and white defendants, is substantially lower than the FPR in the Update, Random, and No Advice treatments, but at the cost of higher FNR. Specifically, the average FPR in the Learned treatment is 31.9\% and 13.2\% for black and white defendants, respectively, which is significantly lower than the FPR of 45.6\% and 29.7\%, respectively, obtained in the Update treatment ($p<10^{-5}$). The FNR in the Learned treatment is 49.8\% and 54.8\% for black and white defendants, respectively, which is higher than the FNR of 38.6\% and 51.3\%, respectively, in the Update treatment, but the difference is statistically significant only for black defendants ($p<10^{-5}$). A comparison of the Learned treatment with the risk-assessment algorithm shows that the FPR for black defendants is significantly lower in the Learned treatment (paired t-test, $p<0.01$) but is comparable to the FPR of the algorithm for white defendants ($p=0.34$), and that the FNR for black defendants is significantly higher in the Learned treatment ($p<10^{-5}$) but is comparable to the FNR of the algorithm for white defendants ($p=0.78$).

Next, we look at the disparity in FPR and FNR between black and white defendants in the Learned treatment. A comparison of the Learned and Update treatments shows significantly lower FNR disparity in the Learned treatment ($p<0.01$). The FPR disparity in Learned is higher than that in Update, but the difference is not statistically significant ($p=0.07$). A comparison of the Learned treatment with the risk-assessment algorithm shows that both the FPR and FNR disparities are significantly lower in the Learned treatment than those of the algorithmic predictions (paired t-test, $p<10^{-5}$). Thus, in terms of FPR and FNR disparities, Learned has a significant advantage compared with the algorithm, and is comparable to or better than the Update treatment.

It can be seen in Figure 5 in the main text that  
according to both FPR and FNR, all treatments have an error that is biased to the same direction which gives harsher predictions for black defendants. 
We quantify this discrimination by defining the  ``classification disparity''  (as described in the main text)  
which weighs utility gaps between black and white defendants that are in favor of white defendants.
Figure \ref{fig:disparity} shows the classification disparity for all experimental treatments and for the risk-assessment algorithm. Interestingly, the algorithm has the highest bias although it did not directly observe race whereas  humans did. Learned, Omniscient, and Update have intermediate classification disparity values, with an advantage to Learned and Omniscient. No Advice and Random have the least classification disparity. 

Figure \ref{fig:tradeoff} shows that when considering the accuracy-fairness tradeoff, Learned and Omniscient Pareto dominate the Update treatment. Omniscient Pareto dominates the algorithm as well. Learned is comparable to the algorithm by accuracy performance (the performance ranking depends on the measure used, as discussed in the main text), but is superior in terms of disparity. These results suggest that the informed advising policies (the Learned and Omniscient treatments) manage to balance the impact of algorithmic advice on human predictions between gaining from the high performance of the algorithm, while moderating its racial disparity.

Finally, we analyze interaction disparity according to the two measures studied in \cite{benCSCW}, to evaluate whether participants responded to the risk assessment in a racially biased manner. In \cite{benCSCW}, every experimental treatment exhibited disparate interactions, including the Update treatment (which is identical to the Update treatment in our experiment) that yielded the smallest disparity. 
Our experiment replicates the results for the ``influence disparity'' measure for the Update treatment. However, in the Learned and Omniscient treatments this influence disparity was eliminated.
Specifically, similar to \cite{benCSCW}, in our Update treatment when the risk assessment was higher than the human's initial prediction, its influence to increase risk in predictions about black defendants was significantly larger than in predictions about white defendants (influence of 0.35 vs. 0.27 for black and white defendants, respectively, paired t-test $p<0.02$), but when the risk assessment was lower than the human's initial prediction, the inverse pattern emerged (though it did not reach statistical significance: influence of 0.28 vs. 0.31 for black and white defendants, respectively, paired t-test $p=0.059$). By contrast, this bias was not observed in the Learned and Omniscient treatments, which as discussed had substantially higher responsiveness to the algorithmic advice. 
Second, the ``deviation disparity'' observed in \cite{benCSCW} was not observed in our experiment, including our Update treatment. Specifically, in \cite{benCSCW} the results show that participants on average deviated positively (toward higher risk) for black defendants and negatively (toward lower risk) for white defendants. This effect was not observed in our experiment, and in fact participants on average deviated positively both for black and white defendants, and the deviation was larger for white defendants (though all differences were small).